\pdfoutput=1
\documentclass[journal]{IEEEtai}
\usepackage[colorlinks,urlcolor=blue,linkcolor=blue,citecolor=blue]{hyperref}
\usepackage{color,array}
\usepackage{mathtools}
\usepackage{graphicx}
\usepackage{subfigure}
\usepackage{algorithm,algorithmic}
\usepackage{graphicx}
\usepackage{caption}
\usepackage{cite}
\usepackage{amsmath,amssymb,amsfonts,mathtools}
\usepackage{amsthm}
\usepackage{algorithmic}
\usepackage{textcomp}
\usepackage{xcolor}
\usepackage{breqn}
\usepackage{amsthm}  
\newtheorem{thm}{Theorem}
\newtheorem{lem}{Lemma}
\newtheorem{define}{Definition}
\newtheorem{inequality}{Inequality}
\newcommand{\emu}{E_{i}^{\mu}}
\newcommand{\tmu}{E_{i}^{\theta}}
\newcommand{\eb}{\mathbb{E}}

\begin{document}
\title{Thompson Sampling with Virtual Helping Agents} 

\author{Kartik Anand Pant, Amod Hegde, and K. V. Srinivas, \IEEEmembership{Member, IEEE}
\thanks{This paragraph of the first footnote will contain the date on which you submitted your paper for review. It will also contain support information, including sponsor and financial support acknowledgment.}

\thanks{Kartik Anand Pant is with Purdue University, USA. This work was done while he was student at the Electronics Engineering Department, IIT (BHU), Varanasi, India (e-mail: kpant@purdue.edu).}

\thanks{Amod Hegde is with Stanford University, USA. This work was done while he was student at the Electronics Engineering Department, IIT (BHU), Varanasi, India (e-mail: amod96@stanford.edu).}

\thanks{K. V. Srinivas is with Motorola Mobility,  Canada. This work was done while he was working at the Electronics Engineering Department, IIT (BHU), Varanasi, India (e-mail: kvsrinivas@ieee.org).}

\thanks{This paragraph will include the Associate Editor who handled your paper.}}

\markboth{Journal of IEEE Transactions on Artificial Intelligence, Vol. 00, No. 0, Month 2020}
{Pant \MakeLowercase{\textit{et al.}}: Thompson Sampling with Virtual Helping Agents}

\maketitle
\begin{abstract}
We address the problem of online sequential decision making, i.e., balancing the trade-off between exploiting the current knowledge to maximize immediate performance and exploring the new information to gain long-term benefits using the multi-armed bandit framework. Thompson sampling is one of the heuristics for choosing actions that address this exploration-exploitation dilemma. We first propose a general framework that helps heuristically tune the exploration versus exploitation trade-off in Thompson sampling using multiple samples from the posterior distribution. Utilizing this framework, we propose two algorithms for the multi-armed bandit problem and provide theoretical bounds on the cumulative regret. Next, we demonstrate the empirical improvement in the cumulative regret performance of the proposed algorithm over Thompson Sampling. We also show the effectiveness of the proposed algorithm on real-world datasets. Contrary to the existing methods, our framework provides a mechanism to vary the amount of exploration/ exploitation based on the task at hand. Towards this end, we extend our framework for two additional problems, i.e., best arm identification and time-sensitive learning in bandits and compare our algorithm with existing methods. 
\end{abstract}

\begin{IEEEImpStatement}
The Multi-Armed Bandit problem has been extensively studied in the last decade. It has wide-ranging applications from clinical trials to product assortment. Recently, it has been utilized in web advertisements and recommendation systems. Thompson sampling (TS) provides a simple heuristic solution to the MAB problem with sub-linear regret bounds. However, TS doesn't provide control over the amount of exploration and exploitation executed by the policy. In this letter, we provide an extension to TS allowing variation in the exploration and exploitation in TS with strong theoretical guarantees. We believe that our framework can be easily augmented with the existing TS based solution and enhance its performance. Our work finds its application even in scenarios where the objective is to find satisficing (nearly optimal) actions on a short horizon or best actions over a long time horizon.             
\end{IEEEImpStatement}


\section{Introduction}
\IEEEPARstart{I}{n} a stochastic multi-armed bandit (MAB) setting, an agent 
faces the problem of sequential decision making in the face of uncertainty. 
At each time step, the agent takes an action from a set of actions and each action produces a reward drawn from an underlying, {\em{fixed but unknown}}, distribution associated with that action. As the agent observes the reward at each time step, she learns about the underlying reward distributions and  tries to optimize her long-term performance. 
The agent faces the dilemma of {\em{exploiting}} the already acquired knowledge to maximize her immediate rewards or {\em{exploring}} actions from which few/no observations have been made to acquire more knowledge for potential future gains while facing the risk of immediate loss.   

Various algorithms have been proposed to solve the exploitation-exploration dilemma in the stochastic MAB problem. They include simple heuristics such as greedy and $\epsilon$-greedy algorithms~\cite{eGreedy}, computationally intensive approaches such as Gittins indices~\cite{Gittins}, and the Upper Confidence Bound (UCB) family of algorithms which offer low computational cost and strong theoretical guarantees on the performance  \cite{LaiRobbins1985}, \cite{Auer2002}, \cite{Audibert2009}, \cite{Garivier2011}, \cite{Maillard2011}, \cite{KaufmannUCB}.  

Thompson proposed a simple heuristic for the stochastic MAB with Bernoulli rewards~\cite{Thompson}. Starting with a prior distribution over the unknown parameters of the reward distribution of each action, the algorithm updates the posterior distributions as the actions are played. At each time step, an action is chosen according to its posterior probability of being the best action. This algorithm is known as {\em{Thompson sampling}} (TS) (and also as {\em{posterior sampling}}, {\em{probability matching}}) and has attracted a lot of attention in recent times. While~\cite{Granmo2010}, \cite{Scott}, \cite{Graepel}, \cite{Chapelle}, \cite{May2011} presented empirical studies showing excellent performance of TS in comparison with other state-of-the-art algorithms along with  some weak theoretical guarantees of TS, \cite{Agrawal2012}, \cite{Agrawal2013a}, \cite{KauffmanTS}, \cite{Gopalan}, \cite{RussoInfTh} and \cite{Agrawal2013J} have presented rigorous theoretical analysis establishing tight bounds on the regret performance of TS. 

In this paper, we present a modified TS algorithm, referred to as {\em{Thompson Sampling with Virtual Helping Agents and Combining}} (TS-VHA-$\mathsf{C}$). The real (or, primary) agent playing the MAB game is assisted by $N-1 > 0$ virtual helping agents, with each agent 
generating an independent sample from the posterior distribution of each arm;
All the $N$ samples ($N-1$ samples generated by the $N-1$ virtual helping agents and the one generated by the primary agent), corresponding to each arm, are processed using a combiner and 
which arm to play next is decided based on the values of the combined samples. Here, we propose two linear combiners $\mathsf{C1}$ and $\mathsf{C2}$. Compared to the (conventional) TS, $\mathsf{C1}$ increases the exploitation at the expense of exploration and $\mathsf{C2}$ increases the exploration at the expense of exploitation. 
    
Importantly, our work may be considered as a framework for varying exploration vs. exploitation for Thompson sampling, by choosing the number of virtual helping agents and the type of combiner, enabling us to achieve a better regret performance (compared to TS) for some of the MAB problems. It is to be noted that one can design other combiners that  achieve a different exploitation-exploration tradeoff. 

Rest of the paper is organized as follows. After introducing the details of stochastic MAB problem and the Thompson sampling in Section \ref{sec:MAB}, we present the TS-VHA algorithm in Section \ref{sec:TSVHA}. Section \ref{sec:analysis} states the main theoretical results that we present and the corresponding proofs. In Section \ref{sec:sims}, we present  simulation results to substantiate our theoretical results and Section \ref{sec:conclusion} concludes the paper.

\section{The Stochastic Multi-armed Bandit Problem}
\label{sec:MAB}
Consider an agent faced with a stochastic MAB problem. Given a slot machine with $K$ arms, the agent has to choose an arm to play at each time step $t\in \mathbb{Z}_{>0}$. The real-valued reward produced by each arm, when played, is a random variable whose distribution is {\em{fixed}} but {\em{unknown}} with a {\em{finite}} support over $[0,1]$. The rewards obtained by playing an arm repeatedly are independent and identically distributed (i.i.d) and are independent of the plays of the other arms. The agent has to decide which arm to play at each time $t$, based on its observations of the past $t-1$ plays and their outcomes, to maximize the {\em{expected total reward}} at time $T$, a widely used performance metric in the stochastic MAB setting. The set of arms can also be referred to as the set of actions and playing arm $i$ is equivalent to choosing action $i$.

Denoting the (unknown) expected reward of arm $i$ with $\mu_i$ and the index of the arm played at time $t$ with $i(t)$, the expected total reward at time $T$ is given by $\mathbb{E}\left[\sum_{t=1}^T \mu_{i(t)}\right]$. An equivalent (and convenient) metric to work with is the {\em{expected total regret}}, given by   
\begin{equation}
    \mathbb{E}[R(T)] = \mathbb{E}\left[\sum_{t=1}^{T}\mu^{*}-\mu_{i(t)}\right],
\end{equation}
where 
$\mu^{\ast} \coloneqq \max_i \mu_i$ and the expectation is over the random choices of arms played by the algorithm. 

\subsection{Thompson Sampling}

As stated before, Thompson sampling takes a Bayesian approach. It starts by assuming an independent prior belief $P(\tilde{\mu}_i)$ over the expected reward of each arm $i$ and a likelihood function $P(r \mid \tilde{\mu}_i)$ representing the probability of observing reward $r$ upon playing arm $i$. When an arm $i$ is played, its posterior is updated based on the observed reward $r$ using the Bayes rule: $P(\tilde{\mu}_i \mid r) \propto P(r \mid \tilde{\mu}_i) P(\tilde{\mu}_i)$. At each time $t$, an arm is played according to its posterior probability of having the highest mean reward; In practice, this is done by simply drawing a sample from the posterior distribution of each arm and playing the arm that produces the largest sample. Algorithm \ref{alg:TS} presents the Thompson sampling.  
\begin{algorithm}[t]
\caption{Thompson Sampling (TS)}
\begin{algorithmic}
\label{alg:TS}
\renewcommand{\algorithmicrequire}{\textbf{Input:}}
\REQUIRE $K$, priors $P(\hat{\mu}_i)$, likelihood $P(r\mid\hat{\mu}_i),~i=1, \ldots, K$.   
\renewcommand{\algorithmicforall}{\textbf{for each}}
\FORALL{$t = 1, 2, \ldots$}
\STATE {\bf{Sample:}} 
\STATE \hspace{0mm}Draw $\theta_i(t) \sim P(\hat{\mu}_i), i=1,\ldots,K$
\STATE{\bf{Select action:}} 
\STATE \hspace{0mm}Play arm  $i(t) = \arg \max_i \theta_i(t)$ and observe its reward $r(t)$
\STATE{\bf{Update distribution:}} 
\STATE \hspace{0mm}$P(\hat{\mu}_{i(t)}) \leftarrow P(\hat{\mu}_{i(t)} \mid r(t))$, \\ where  $P(\hat{\mu}_{i(t)} \mid r(t)) \propto P(r(t) \mid \hat{\mu}_{i(t)}) P(\hat{\mu}_{i(t)})$
\ENDFOR
\end{algorithmic} 
\end{algorithm}

\section{Thompson Sampling with Virtual Helping Agents}
\label{sec:TSVHA}
Thompson sampling has three essential steps. First, the agent {\em{draws a sample}} from the posterior distribution of the expected reward of each arm, which acts as an estimate of the arm's expected reward. Next, the agent {\em{selects the arm}} with the largest sample 
and observes a reward. Finally, the agent {\em{updates}} the posterior distribution of the expected reward of the selected arm based on the observed reward.

Our proposed algorithm introduces two significant changes to the Thompson sampling. 
First, we modify the sampling step of TS by employing $N-1$ {\em{virtual helping agents}}. Let $\mathcal{K}=\{1,2,\ldots,K\}$ represent the set of arms and let $\mathcal{A}=\{1\} \cup \{2,\ldots,N\} = \{1, 2,\ldots,N\}$ denote the set of {\em{all}} agents, containing the real agent and the $N-1$ virtual helping agents. At every time step $t$, all the $N$ agents perform the sampling activity; i.e., every agent $n\in \mathcal{A}$ draws a sample, {\em{independently}}, from $P(\hat{\mu}_i)$, the posterior distribution of the expected reward of $i^{\text{th}}$ arm, $\forall i \in \mathcal{K}$. At the end of the sampling step, each agent $n\in\mathcal{A}$ 
has $K$ samples $\theta_{i,n}(t), i = 1, \ldots, K$, where $\theta_{i,n}(t)$ is the sample drawn by agent $n$ from the posterior distribution of arm $i$ at time $t$. Note that the sampling activity (for generating the samples $\theta_{i,n}(t), i = 1, \ldots, K, n = 1, \ldots, N$) is {\em{independent across the agents and across the arms}}. 

Next, for each arm $i\in\mathcal{K}$, we combine the samples $\theta_{i,n}(t), n=1, \ldots, N$, 
using a combiner $f:\mathbb{R}^{1 \times N} \rightarrow \mathbb{R}$ to arrive at the final \textit{combined} estimate of the expected reward $\theta_{i}(t)$ of the $i^{\text{th}}$ arm. After the combining step, like in TS, we select the arm with the largest {\em{combined}} sample, observe the reward and update the posterior of the selected arm based on the observed reward. Note that the posterior update is same as that of the TS. 

\begin{algorithm}[t]
\caption{TS with Virtual Helping Agents (TS-VHA)}
\begin{algorithmic}
\label{alg:TSVHA}
\renewcommand{\algorithmicrequire}{\textbf{Input:}}
\REQUIRE $K$, $N$, priors $P(\hat{\mu}_i)$, likelihood $P(r\mid\hat{\mu}_i),~i=1, \ldots, K$,  Combiner $f$.   
\renewcommand{\algorithmicforall}{\textbf{for each}}
\FORALL{$t = 1, 2, \ldots$}
\STATE {\bf{Sample:}} 
\FORALL{$n = 1, 2, \ldots,N$}
\STATE \hspace{0mm}Draw $\theta_{i,n}(t) \sim P(\hat{\mu}_i), \forall i \in \mathcal{K}$
\ENDFOR
\STATE {\bf{Combine:}}
\STATE \hspace{0mm}$\theta_{i}(t)  = f(\theta_{i,1}(t), \ldots, \theta_{i,N}(t)), \forall i \in \mathcal{K}$
\STATE{\bf{Select action:}} 
\STATE \hspace{0mm}Play arm  $i(t) = \arg \max_i \theta_i(t)$ and observe its reward $r(t)$
\STATE{\bf{Update distribution:}} 
\STATE \hspace{0mm}$P(\hat{\mu}_{i(t)}) \leftarrow P(\hat{\mu}_{i(t)} \mid r(t))$, \\ where  $P(\hat{\mu}_{i(t)} \mid r(t)) \propto P(r(t) \mid \hat{\mu}_{i(t)}) P(\hat{\mu}_{i(t)})$
\ENDFOR
\end{algorithmic} 
\end{algorithm}

Algorithm \ref{alg:TSVHA} details the proposed Thompson sampling with virtual helping agents (TS-VHA). 
With no virtual agents (i.e., $N-1=0$) and identity function as the combiner (i.e., $f(\theta_i)=\theta_i)$, 
TS-VHA reduces to TS. In other words, TS-VHA can be interpreted as TS with $N-1$ virtual helping agents and a combiner. The  $N-1$ virtual agents provide the real agent (who is actually trying to solve the MAB problem) with additional samples from the posterior to help her in manipulating the exploitation vs. exploration and deciding which arm to play at each time step.The agents are {\em{virtual}} as they do {\em{not}} really play the arms.   Compared to TS, the additional cost of TS-VHA is in generating $N-1$ additional samples and processing them through the combiner function. 

In this work, we propose two combiners, $\mathsf{C1}$ and $\mathsf{C2}$, which enable us to increase the exploitation (at the cost of exploration) and exploration (at the cost of exploitation), respectively. 
Both the combiners are {\em{linear}}, having the generic form given below.  
\begin{equation}
    f(\theta_{i,1}(t), \ldots, \theta_{i,N}(t)) = \sum_{n=1}^{N} c_n \theta_{i,n}, \forall i \in \mathcal{K}. 
    \label{comb}
\end{equation}
Combiners $\mathsf{C1}$ and $\mathsf{C2}$ differ only in the choice of the coefficients $c_n$ as described in the following. 

\subsection{Combiner $\mathsf{C1}$: Increasing Exploitation}
\label{sec:C1}

Combiner $\mathsf{C1}$ is given by 
\begin{equation}
\label{comb1}
    c_n = \frac{1}{N}, \forall n \in \mathcal{A} 
\end{equation} 
Employing combiner $\mathsf{C1}$ in TS-VHA (which will be referred to as TS-VHA-$\mathsf{C1}$ now onward) leads to higher \textit{exploitation} as compared to TS. Observe that, for any arm $i$, the variance of the distribution of $\theta_{i}(t)$ is $\frac{1}{N}$ times that of the distribution of $\theta_{i,n}(t), n=1,\ldots,N$, whereas the mean remains the same. Thus, for each arm, the variance of the distribution of the combined sample is lower compared to the variance of the posterior distribution of that particular arm. Thus, TS-VHA-$\mathsf{C1}$ places more confidence on its empirical estimates $\hat{\mu}_i(t)$, resulting in increased exploitation and lower exploration when compared with TS.
As $N\rightarrow \infty$, $\theta_i(t) \rightarrow \hat{\mu}_i(t)$ and TS-VHA-$\mathsf{C1}$ emulates greedy decision making.
\subsection{Combiner $\mathsf{C2}$: Increasing Exploration}
\label{sec:C2}
Combiner $\mathsf{C2}$ is for increasing the variance of the distribution of $\theta_i(t)$, the combined sample for arm $i$, compared to the posterior distribution of arm $i$ and is given by the following set of coefficients $c_n$:
\begin{flalign}
    & When~N~is~an~even~integer, \notag \\ 
    &c_n = \frac{1}{N}+\left(\sqrt{\frac{N^2+1}{N}}\right)^{n+1}, \ n=1, \ldots, N.  \label{comb2even}\\
    & When~N~is~an~odd~integer, \notag \\
    &c_n = \left\{ \begin{array}{l l}  \frac{1}{N}+\left(\sqrt{\frac{N+1}{N}}\right)^{n+1}, & n=1, \ldots, N-1, \label{comb2odd} \\
    \frac{1}{n}, & n=N. 
    \end{array} \right.
\end{flalign}

Observe that, for any arm $i$, the variance of the distribution of $\theta_{i}(t)$ is $N$ times that of the distribution of $\theta_{i,n}(t), n=1,\ldots,N$, whereas the mean remains the same. Due to this increase in the variance, TS-VHA with $\mathsf{C2}$ as its combiner (which will be referred to as TS-VHA-$\mathsf{C2}$ hereafter) places less confidence on its empirical estimates $\hat{\mu}_i(t)$ and leads to higher  {\em{exploration}} as compared to TS. Increasing the number of agents in this case makes TS-VHA-$\mathsf{C2}$ to over-explore.

Though our focus in this work is on the two linear combiners $\mathsf{C1}$ and $\mathsf{C2}$ (given by (\ref{comb}), (\ref{comb1}), (\ref{comb2even}), (\ref{comb2odd})), note that TS-VHA provides a {\em{generic framework}} to manipulate exploitation vs. exploration. 
One can design other forms of combiners with desired exploitation-exploration tradeoff for a wide-range of MAB problems for which TS can be applied. To highlight this point, we present $\mathsf{C3}$, a third combiner. 

\subsection{Combiner $\mathsf{C3}$: Dynamic Exploitation}
\label{sec:C3}

$\mathsf{C3}$ is a non-linear combiner that computes the combined sample $\theta_i(t)$ for each arm $i\in \mathcal{K}$ as   
\begin{equation}
\label{eq:maxop}
    \theta_i(t) = \max\left(\sum_{n=1}^{N(t)} \frac{1}{N(t)}\theta_{i,n}(t),\min_{j \in \mathcal{K}}(\hat{\mu}_j(t))\right), \forall i \in \mathcal{K}
\end{equation}
where $\hat{\mu}_j(t)$ is the observed empirical mean reward of arm $i$ at time $t$, $\theta_{i,n}(t)$ is the sample generated by agent $n$ for arm $i$ at time $t$ and $N(t)$ is the total number of agents (among which $N-1$ are the virtual helping agents) which is dynamically determined at each time step $t$ as,
\begin{equation}
\label{eq:n_t}
    N(t) = \left \lfloor{\max(1, t\cdot\tilde{\Delta})}\right \rfloor
\end{equation}
where $\tilde{\Delta}=\hat{\mu}^{(1)}(t)-\hat{\mu}^{(2)}(t)$, and $\hat{\mu}^{(1)}(t)$, $\hat{\mu}^{(2)}(t)$ are the largest and the second-largest values, respectively, in the set $\{\hat{\mu}_1(t), \ldots, \hat{\mu}_K(t)\}$.

 
In \eqref{eq:maxop}, the term $\sum_{n=1}^{N(t)} \frac{1}{N(t)}\theta_{i,n}(t)$ is similar to combiner $\mathsf{C1}$ with $N(t)$ number of agents. Note that, here, the number of agents $N(t)$ is a function of time unlike in $\mathsf{C1}$. As discussed previously, the value of $N(t)$ commands the exploitation-exploration trade-off. The expression for $N(t)$ in \eqref{eq:n_t} is based on the following intuition:
\begin{itemize}
    \item {With time, we expect our best empirical arm to be the optimal arm with increasing confidence. Thus, tuning TS to increase exploitation with time could reduce the regret incurred at later time steps and improve its performance.}
    
    \item {If the difference in the empirical means of the top two candidate arms $\tilde{\Delta}$ is high, it may suggest that the best empirical arm is indeed the optimal arm. Increasing exploitation in this case may help reduce the cumulative regret on average. On the other hand, if $\tilde{\Delta}$ is low, we may have to explore more to better discern the optimal arm. Hence, we set $N(t) \propto \tilde{\Delta}$.   }   
    
\end{itemize}

Based on the above intuition{\footnote{We would like to emphasize that these points are only our intuition and we do not have a mathematical justification}}, we hypothesize that the number of agents to be deployed should be dependent on time $t$ and $\hat{\mu}^1(t) - \hat{\mu}^2(t)$. Next, inspired from \cite{May2011} we apply a max operation~\eqref{eq:maxop}. Note that, unlike OBS, we take the \textit{maximum} of the averaged estimate and the \textit{minimum} among the empirical means of all arms. We observe that this step yields superior empirical results.

In Section~\ref{sec:sims}, we provide  simulation studies that prove the effectiveness of TS-VHA, for all the three combiners $\mathsf{C1}$, $\mathsf{C2}$ and $\mathsf{C3}$, on Bernoulli bandits and Gaussian bandits. However, due to the analytical tractability of Gaussian distribution, we focus on Gaussian bandits while analyzing the regret performance of TS-VHA. We provide the mathematical regret analysis only for combiners $\mathsf{C1}$ and $\mathsf{C2}$ and leave the analysis of the non-linear combiner $\mathsf{C3}$ for future work. 

\subsection{Gaussian Bandits}
In the rest of the paper, we mainly focus on  stochastic multi-armed bandits {where the likelihood of the reward distributions are Gaussian; To be precise, the likelihood of the reward of arm $i\in \mathcal{K}$ is  Gaussian distributed with mean $\mu_i$ ({\em{unknown a priori}}) and unit variance}{\footnote{As in~\cite{Agrawal2013a}, we consider the single-parameter model where only the mean of the reward distribution is unknown. We do not consider the two-parameter model where both mean and variance of the reward distribution are unknown.}}.
Equivalently, the likelihood of $r_i(t)$, reward from arm $i$ at time $t$, given parameter $\mu_i(t)$, is given by $\mathcal{N}(\mu_i(t),1)$. 
Denote the arm played at time $t$ with $i(t)$ and the number of plays of arm $i$ until (and including) $t-1$ with $k_i(t)$. Define $\hat{\mu}_i(t) \coloneqq \frac{\sum_{\tau=1:i(\tau)=i}^{t-1}r_i(\tau)}{k_i(t)+1}$, and $\hat{\mu}_i(1) \coloneqq 0$.  
With the Gaussian likelihood, it is convenient to use Gaussian priors. Consider $\mathcal{N}\left(\hat{\mu}_i(t), \frac{1}{k_i(t)+1}\right)$ as the prior for $\mu_i$ at time $t$. When arm $i$ is played at time $t$, the posterior distribution for $\mu_i$, by applying Bayes rule, turns out to be $\mathcal{N}\left(\hat{\mu}_i(t+1),\frac{1}{k_i(t+1)+1}\right)$. 

By using the Gaussian priors and likelihoods in Algorithm~\ref{alg:TS}, TS can be employed for Gaussian bandits and is referred to as {\em{TS using Gaussian priors}}.   
\subsection{TS-VHA using Gaussian Priors} 
For a Gaussian bandit, we can apply TS-VHA by using Gaussian priors and Gaussian likelihoods in Algorithm~\ref{alg:TSVHA} resulting in Algorithm~\ref{alg:TSVHA_G}. The additional step of generating multiple samples and combining them in TS-VHA may alter the distribution of $\theta_i(t)$ in Algorithm~\ref{alg:TSVHA}. However, with Gaussian distributions and a linear combiner $f$ (such as the one given by (\ref{comb})), the distribution of $\theta_i(t)$ remains Gaussian.

\begin{algorithm}[t]
\caption{TS-VHA using Gaussian Priors}
\begin{algorithmic}
\label{alg:TSVHA_G}
\renewcommand{\algorithmicrequire}{\textbf{Input:}}
\REQUIRE $\mathcal{K}, N,  Set \  \mu_i(1) = 0 \: \forall \: i \in $
priors $P(\hat{\mu}_i)$, likelihood $P(r\mid\hat{\mu}_i),~i=1, \ldots, K$,  Combiner $f$.   
\renewcommand{\algorithmicforall}{\textbf{for each}}
\FORALL{$t = 1, 2, \ldots$}
\STATE {\bf{Sample:}} 
\FORALL{$n = 1, 2, \ldots,N$}
\STATE \hspace{0mm}Draw $\theta_{i,n}(t) \sim P(\hat{\mu}_i), \forall i \in \mathcal{K}$
\ENDFOR
\STATE {\bf{Combine:}}
\STATE \hspace{0mm}$\theta_{i}(t)  = f(\theta_{i,1}(t), \ldots, \theta_{i,N}(t)), \forall i \in \mathcal{K}$
\STATE{\bf{Select action:}} 
\STATE \hspace{0mm}Play arm  $i(t) = \arg \max_i \theta_i(t)$ and observe its reward $r(t)$
\STATE{\bf{Update distribution:}} 
\STATE \hspace{0mm}$P(\hat{\mu}_{i(t)}) \leftarrow P(\hat{\mu}_{i(t)} \mid r(t))$, \\ where  $P(\hat{\mu}_{i(t)} \mid r(t)) \propto P(r(t) \mid \hat{\mu}_{i(t)}) P(\hat{\mu}_{i(t)})$
\ENDFOR
\end{algorithmic} 
\end{algorithm}

In the next Section, we bound the finite time expected regret of TS-VHA-$\mathsf{C1}$ and TS-VHA-$\mathsf{C2}$ for Gaussian bandits. 
Note that we have investigated the performance of TS-VHA-$\mathsf{C3}$ only through simulation experiments, presented in Section \ref{sec:sims}. Understanding the theoretical implications of $\mathsf{C3}$ would be interesting and we will consider it in our future work. 

\section{Regret Analysis}
\label{sec:analysis}

For the finite time regret analysis presented in this section, we consider employing TS-VHA with $N-1 > 0$ virtual helping agents for Gaussian bandits with reward distribution being finite support over [0, 1]. 
When we chose $\mathsf{C1}$ as the combiner, the variance of the combined sample $\theta_i(t), \forall i \in \mathcal{K}$, gets scaled by $1/N$ compared to the variance of the posterior distribution of arm $i, \forall i \in \mathcal{K}$. With combiner $\mathsf{C2}$, variance of $\theta_i(t), \forall i \in \mathcal{K}$, gets scaled by $N$ compared to the variance of the posterior distribution of arm $i, \forall i \in \mathcal{K}$. 

In both the cases, the mean of the combined sample for each arm is equal to the mean of the posterior distribution of the corresponding arm.  
Equivalently, at each time step $t$, TS-VHA-$\mathsf{C1}$ results in scaling the variance of $\theta_i(t), \forall i \in \mathcal{K}$ by a factor $1/N$ and TS-VHA-$\mathsf{C2}$ scales the variance of $\theta_i(t), \forall i \in \mathcal{K}$ by $N$, {\em{when compared to Thompson sampling}}. 

To unify the regret analysis of TS-VHA-$\mathsf{C1}$ and TS-VHA-$\mathsf{C2}$, we introduce $1/\gamma$ as the factor that determines the variance scaling. 
Thus, $\gamma > 1$ corresponds to TS-VHA-$\mathsf{C1}$,  $\gamma\in(0,1)$ corresponds to TS-VHA-$\mathsf{C2}$. 

\begin{thm}
\label{theorem1}
For the $K$-armed stochastic bandit problem, Thompson sampling with virtual helping agents using Gaussian priors and with variance scaling factor $\gamma$ has expected regret   
at time $T \geq K$. \\  
$For\ \gamma\ \in (0,4),$
\begin{multline}
 \mathbb{E}[R(T)] 
   \leq \sum_{i=2}^{K} \left(c^1_i \ln{T\Delta_i^2} + f_i^1(\beta,\gamma,\epsilon)\Delta_i+ \frac{9.5}{\Delta_i}\right)  
\end{multline}
$For\ \gamma\ \geq 4,$
\begin{multline}
\mathbb{E}[R(T)] 
   \leq \sum_{i=2}^{K}(c^1_i \ln{T\Delta_i^2} +    
   \\c'\left(\frac{T^{1+\epsilon-\frac{2\beta}{\gamma}}-1 }{1+\epsilon-\frac{2\beta}{\gamma}} + g(\epsilon)+1\right)\Delta_i 
   + \frac{9.5}{\Delta_i})
\end{multline}

where {$c_i^1 = \frac{2(H(\beta)+1)\Delta_i}{\gamma (y_i-x_i )^2}, f_i^1(\beta,\gamma,\epsilon) = c'(g(e)+\zeta(\frac{2\beta}{\gamma}-\epsilon) )+1$} and $\beta, \epsilon, y_i, x_i, \Delta_i, c'$ are all constants at time $T \geq K$.  
\end{thm}

\subsection{Proof of Theorem~\ref{theorem1}}
We adopt the notation and definitions from~\cite{Agrawal2013J}  and follow the same methodology as that of~\cite{Agrawal2013J} in analyzing the finite cumulative regret achieved by TS-VHA using Gaussian priors when employed over a $K$ armed Gaussian bandit. Without loss of generality, we assume that $\mu^{\ast}=\mu_1> \arg \max_{i\neq 1} \mu_i$.    

\begin{define}
$i(t)$ denotes the arm played at time $t$, $k_{i}(t)$ denotes the number of plays of arm $i$ until, and including, time $t-1$. $\hat{\mu}_i(t)$ denotes the empirical mean, given by $\hat{\mu}_i(t)=\frac{\sum_{\tau=1:i(\tau)=i}^{t-1}r_i(\tau)}{k_i(t)+1}$, where $r_i(t)$ denotes the reward observed from arm $i$ at time $t$ and $\hat{\mu}_i(t)=0$ when $k_i(t)=0$.
\end{define}
\begin{define}
$\theta_{i,n}(t)$ denotes the $n^{\text{th}}$ sample generated, independently, from $\mathcal{N}\left( \hat{\mu}_i(t), \frac{1}{k_i(t)+1}\right)$, the posterior distribution of arm $i$ at time $t$ and $\theta_{i}(t)=f(\theta_{i,1}, \ldots, \theta_{i,N})$.
\end{define}
\begin{define}
For arm $i=2, \ldots, K$, $x_i$ and $y_i$ denote thresholds such that $\mu_i<x_i<y_i<\mu_1$. 
\end{define}
\begin{define}
For $i=2,\ldots, K$, $E_i^\mu(t)$ is the event $\hat{\mu}_i(t)\leq x_i$ and $E_i^\theta(t)$ is the event $\theta_i(t)\leq y_i$.
\end{define}

\begin{define}
$\mathcal{F}_t = \{i(\tau),r_{i(\tau)}(\tau),\tau=1,2,\ldots, t\}$ is the history of arm play until time $t$, 
where $i(\tau)$ is the arm played at time $\tau$ and $r_{i(\tau)}(\tau)$ is the reward observed from arm $i(\tau)$ at time $\tau$. Define $\mathcal{F}_0=\emptyset$. By definition, $\mathcal{F}_0 \subseteq \mathcal{F}_1 \subseteq \ldots \mathcal{F}_{T-1}$.
\end{define}

\begin{define}
\label{defpit}
Define $p_{i,t}$ as the probability
\begin{equation*}
    p_{i,t} = \text{Pr}(\theta_1(t)>y_i\mid\mathcal{F}_{t-1}).
\end{equation*}
\end{define}

The expected total regret in time $T$ is given by 
\begin{equation}
    \mathbb{E}[R(T)] = \mathbb{E}\left[\sum_{t=1}^{T}(\mu^{*}-\mu_{i(t)})\right]=\sum_{i}\Delta_{i}\mathbb{E}[k_{i}(T)], \label{eq:regret}
\end{equation}
where $\Delta_i=\mu^{\ast}-\mu_i$ and $\mu^{\ast} \coloneqq \max_i \mu_i$. 
In order to bound the expected regret, we need to bound $\mathbb{E}[k_{i}(T)]$ for $i \neq 1$, 

which can be decomposed into three terms as follows: 
\begin{align}
    \mathbb{E}[k_{i}(T)]&=\sum_{t=1}^{T}\text{Pr}(i(t)=i) \label{eq:main1} \\
                        &=\sum_{t=1}^{T}\text{Pr}\left(i(t)=i,E_{i}^{\mu}(t), E_{i}^{\theta}(t)\right)\tag{A}\label{eq:1} \\
                        &\quad+\sum_{t=1}^{T}\text{Pr}\left(i(t)=i,\emu(t), \overline{\tmu(t)}\right)\tag{B}\label{eq:2}\\  &\qquad+\sum_{t=1}^{T}\text{Pr}\left(i(t)=i,\overline{\emu(t)}\right)\tag{C}\label{eq:3}
\end{align}
We will now consider the terms \eqref{eq:1}, \eqref{eq:2} and \eqref{eq:3} individually. 
In the following, we consider $x_i=\mu_i+\frac{\Delta_i}{3}$,  $y_i=\mu_1-\frac{\Delta_i}{3}$ and let $L_i(T) = \frac{2 \ln{T\Delta_i^2}}{\gamma(y_i-x_i)^2}$.
\subsection{Term (\ref{eq:1})}

For $k\geq 1$, let $\tau_{k}$ be the time step at which
the first arm is played for the $k^{\text{th}}$ time, and let $\tau_{0} = 0$.
Then, as shown in~\cite{Agrawal2013J}
(cf. Eqn. (4), Section 2.1 in~\cite{Agrawal2013J}),  
for $i \neq 1$, 
\begin{equation}
\label{eqn:termA1}
    \sum_{t=1}^{T}\text{Pr}\left(i(t)=i,E_{i}^{\mu}(t), E_{i}^{\theta}(t)\right) \leq \sum_{k=0}^{T-1}\eb\left[\frac{(1-p_{i,\tau_{k}+1})}{p_{i,\tau_{k}+1}}\right].
\end{equation}
It is easy to verify that  (\ref{eqn:termA1}) does {\em{not}} get affected by the distribution of the sample $\theta_i(t), \forall i \in \mathcal{K}$  . Hence, (\ref{eqn:termA1})  holds good for TS as well as TS-VHA. 
We utilize (\ref{eqn:termA1}) to prove the following bound on term \eqref{eq:1}.
\begin{lem}
\label{lem:termA}
{\text{For}} $\gamma\in (0,4)$, 
\begin{multline}
    \sum_{t=1}^{T} \text{Pr} ( i(t) = i, E_{i}^{\mu}(t), E_{i}^{\theta}(t))
    \leq H(\beta)L_i(T) + \\ c' \left( g(\epsilon) + \zeta(\frac{2\beta}{\gamma}-\epsilon) \right) + \frac{4}{\Delta_i^2},
\end{multline}
\text{For} $\gamma \geq 4$,
\begin{multline}
    \sum_{t=1}^{T}\text{Pr}( i(t) = i, E_{i}^{\mu}(t), E_{i}^{\theta}(t))
    \leq H(\beta)L_i(T) + \\ c' \left( g(\epsilon) + \frac{T^{1+\epsilon-\frac{2\beta}{\gamma}}-1 }{1+\epsilon-\frac{2\beta}{\gamma}}\right) + \frac{4}{\Delta_i^2},
\end{multline}
\text{where} $\beta \in [1,2)$, $\epsilon > 0$ \text{and} $\zeta$ is the Riemann zeta function.
\end{lem}
\begin{IEEEproof}
Please refer to Appendix~\ref{app:lemma1}.
\end{IEEEproof}

\subsection{Term (\ref{eq:2})}
\begin{lem}
\label{lem:termB}
For $i \neq 1$,
\begin{equation*}
    \sum_{t=1}^{T}Pr\left(i(t)=i,E_{i}^{\mu}(t), \overline{E_{i}^{\theta}(t)}\right) \leq L_i(T)+\frac{1}{\Delta_i^2} 
\end{equation*}
\end{lem}
\begin{IEEEproof}
$\sum_{t=1}^{T}\text{Pr}\left(i(t)=i,E_{i}^{\mu}(t), \overline{E_{i}^{\theta}(t)}\right)$ can be subdivided into two parts based on the values of $k_i(T)$.
\begin{multline}
\sum_{t=1}^{T}\text{Pr}\left(i(t)=i,E_{i}^{\mu}(t), \overline{E_{i}^{\theta}(t)}\right) = \\                \sum_{t=1}^{T}\text{Pr}\left(i(t)=i,E_{i}^{\mu}(t),k_i(T)\leq L_i(T), \overline{E_{i}^{\theta}(t)}\right) \\ + \sum_{t=1}^{T}\text{Pr}\left(i(t)=i,E_{i}^{\mu}(t),k_i(T)> L_i(T), \overline{E_{i}^{\theta}(t)}\right)\label{eqn:termBsplit}
\end{multline}
The first term on the RHS of \eqref{eqn:termBsplit} is bounded by $\mathbb{E}\left[\sum_{t=1}^T I\left(i(t)=i,k_i(t)<L_i(T)\right) \right]$ which is upper bounded  by $L_i(T)$. We now bound the second term on the RHS. 
\begin{multline}
\sum_{t=1}^{T}\text{Pr}\left(i(t)=i,E_{i}^{\mu}(t),k_i(T)> L_i(T), \overline{E_{i}^{\theta}(t)}\right) \\
                \leq \mathbb{E}\left[\sum_{t=1}^T \text{Pr}\left(i(t)=i,\overline{E_{i}^{\theta}(t)}\;\middle|\; k_i(t)>L_i(T),E_{i}^{\mu}(t),\mathcal{F}_{t-1}\right)\right] \\
                \leq \mathbb{E}\left[\sum_{t=1}^T \text{Pr}\left( \theta_i(t)>y_i\;\middle|\; k_i(t)>L_i(T),\hat{\mu}_i(t)\leq x_i,\mathcal{F}_{t-1} \right)\right]
\end{multline}
Note that, $\theta_i(t) \sim \mathcal{N}\left(\hat{\mu}_i(t),\frac{1}{\gamma(k_i(t)+1)}\right)$. 
Let $\nu_i(t) \sim \mathcal{N}\left(x_i,\frac{1}{\gamma(k_i(t)+1)}\right)$. Then, as $\hat{\mu}_i(t)\leq x_i$ 
\begin{multline}
     \text{Pr}\left(\theta_i(t)>y_i|k_i(t)>L_i(T),\hat{\mu}_i(t)\leq x_i,\mathcal{F}_{t-1}\right) \\
     \leq \text{Pr}\left(\nu_i(t)>y_i|k_i(t)>L_i(T),\hat{\mu}_i(t)\leq x_i,\mathcal{F}_{t-1}\right)
\end{multline}
Using \hyperref[ineq3]{Inequality 3}, for any fixed $k_i(t)>L_i(T)=\frac{2\ln(T\Delta_i^2)}{\gamma(y_i-x_i)^2}$,
\begin{align}
     \text{Pr}(\nu_i(t)>y_i) & \leq \frac{1}{2} e^{-\frac{\gamma(k_i(t)+1)(y_i-x_i)^2}{2}} \notag \\     & \leq \frac{1}{2} e^{-\frac{\gamma L_i(T)(y_i-x_i)^2}{2}}\notag \\ & \leq \frac{1}{T \Delta_i^2}
\end{align}
This results in, 
\begin{equation}
    \sum_{t=1}^T \text{Pr}\left( \theta_i(t)>y_i\;\middle|\; k_i(t)>L_i(T),\hat{\mu}_i(t)\leq x_i,\mathcal{F}_{t-1} \right) \leq \frac{1}{\Delta_i^2}, \notag
\end{equation}
bounding the second term on the RHS of (\ref{eqn:termBsplit}) with $\frac{1}{\Delta_i^2}$. 
\end{IEEEproof}

\subsection{Term (\ref{eq:3})}
Term \eqref{eq:3} denotes the probability of pulling the sub-optimal arm $i$ when it is neither well estimated nor well sampled.  
\begin{lem}
\label{lem:termC}
For $i \neq 1$,
\begin{equation*}
    \sum_{t=1}^{T}\text{Pr}\left(i(t)=i,\overline{E_{i}^{\mu}(t)}\right)\leq \frac{1}{d_i(x_i,\mu_i)} \leq \frac{9}{2\Delta_i^2}+1. 
\end{equation*}
\end{lem}
\begin{IEEEproof}
The proof for Lemma~\ref{lem:termC} follows from \cite{Agrawal2013J}. Since, the proof for the Lemma~\ref{lem:termC} doesn't depend on the 
posterior distribution of the arms, the proof provided for Lemma 2.15 in\cite{Agrawal2013J} holds valid as a proof for our Lemma~\ref{lem:termC}. 
\end{IEEEproof}
$\mathbb{E}[k_{i}(T)]$ can be bounded by substituting Lemma~\ref{lem:termA},~\ref{lem:termB} and~\ref{lem:termC} in~\eqref{eq:main1} and using this bound on $\mathbb{E}[k_{i}(T)]$ in \eqref{eq:regret} completes the proof of Theorem \ref{theorem1}. 


\section{Simulation Experiments}
\label{sec:sims}
\begin{figure*}[t]
    \centering
    \subfigure[20 Arms]{%
    \includegraphics[width=16.5pc]{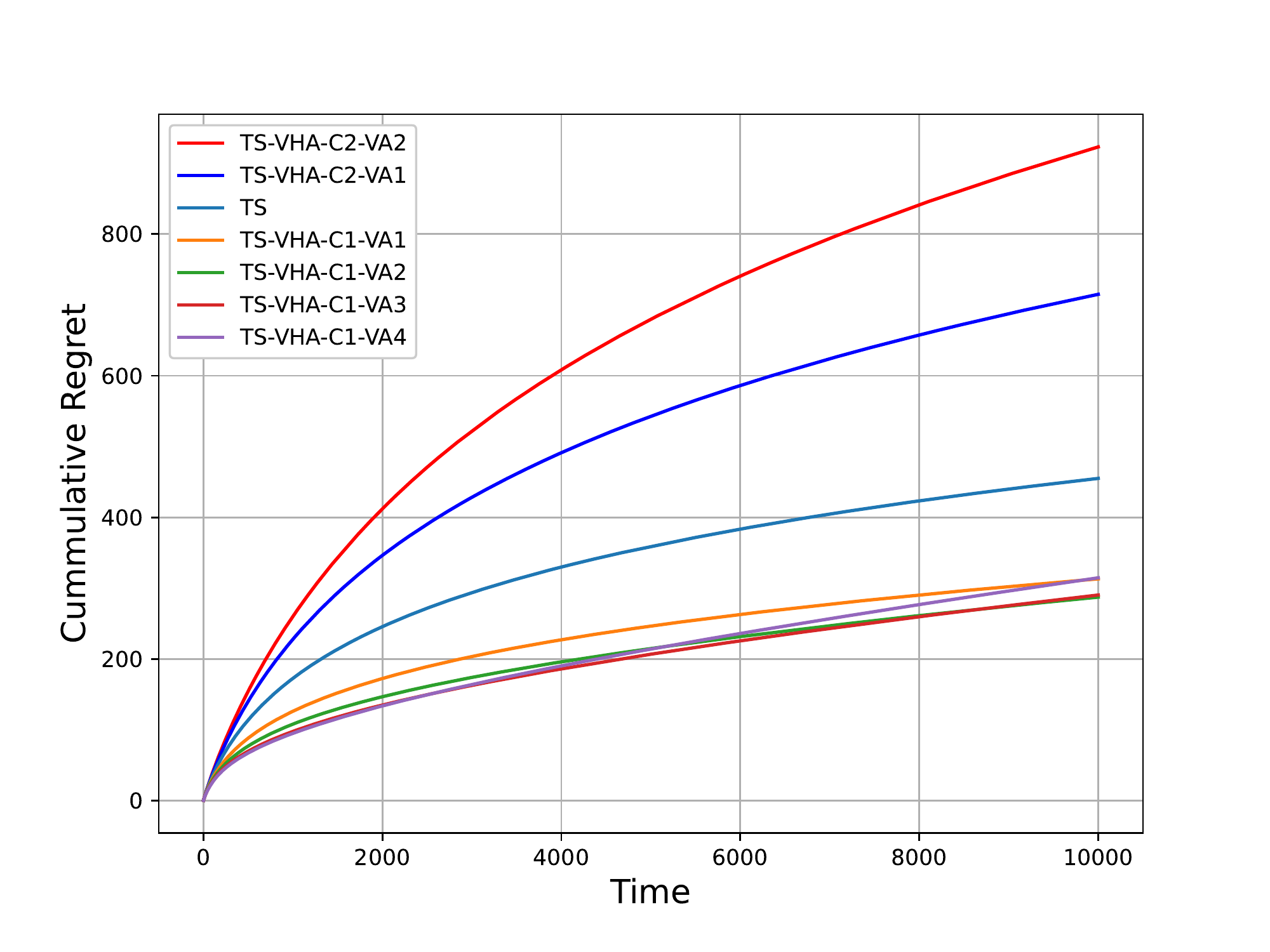}
    \label{fig:RandGsubfig_a}}
    \quad
    \subfigure[200 Arms]{%
    \includegraphics[width=16.5pc]{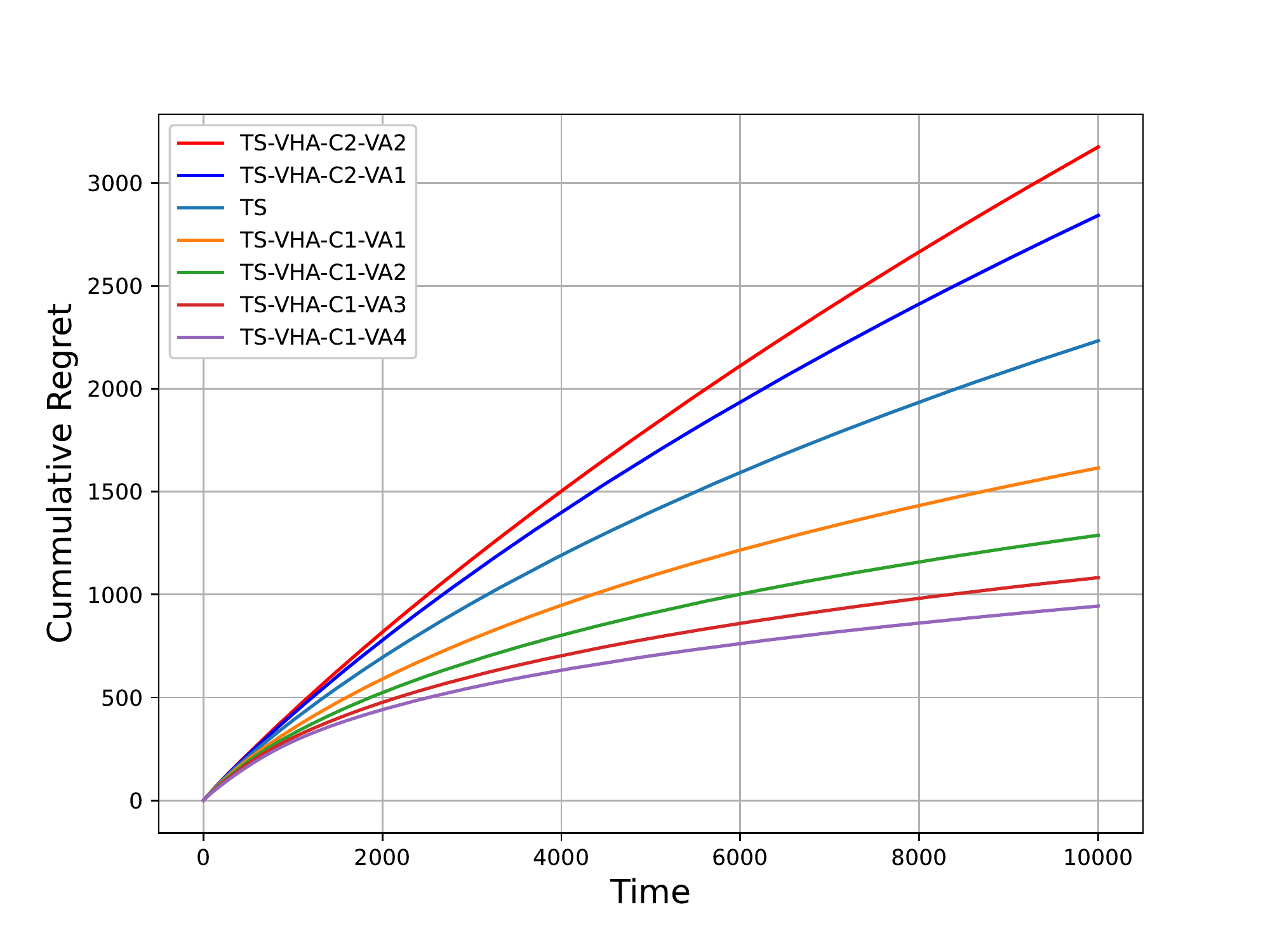}
    \label{fig:RandGsubfig_b}}
    \caption{Gaussian bandit: Cumulative regret comparison of TS with TS-VHA-$\mathsf{C1}$ and TS-VHA-$\mathsf{C2}$.}
    \label{fig:RandG}
\end{figure*}
In this section, we present computational experiments that illustrate the potential benefits of TS-VHA. In the following,  TS-VHA-$\mathsf{C1}$-VA$n$ and TS-VHA-$\mathsf{C2}$-VA$n$ denote TS-VHA with $n$ virtual helping agents, with combiner $\mathsf{C1}$ and $\mathsf{C2}$, respectively. Note that, as per the notation introduced in Section \ref{sec:TSVHA}, $N-1=n$ and TS corresponds to $N=1$ with identity function as the combiner.
{\subsection{Gaussian Bandits}} 
We evaluate the performance of  TS-VHA-$\mathsf{C1}$ and TS-VHA-$\mathsf{C2}$ and compare it with TS for Gaussian bandits. 
First, we consider a 20 armed bandit problem with reward from arm $i$ modeled as $\mathcal{N}(\mu_i,1)$, where the mean reward $\mu_i$ is independently sampled from $\mathcal{U}[0,1]$.
Fig. \ref{fig:RandGsubfig_a} shows the cumulative regret over 10000 time steps, averaged over 1000 independently sampled problem instances. Fig. \ref{fig:RandGsubfig_b} corresponds to a second Gaussian bandit problem with 200 arms, keeping all the other details same as that of the 20 armed bandit. 

For TS-VHA-$\mathsf{C1}$-VA$n$ (TS-VHA-$\mathsf{C2}$-VA$n$), exploitation (exploration) increases with $n$, as compared to TS.   
As can be observed from the plots, increasing exploitation through TS-VHA-$\mathsf{C1}$ improves the regret performance. 
It should be noted that having more exploitation might turn out to be counter-productive. As discussed in Section~\ref{sec:C1}, 
as $n$ grows to a higher value, TS-VHA-$\mathsf{C1}$-VA$n$ starts behaving like the greedy algorithm. Observe that, in Fig.~\ref{fig:RandGsubfig_a}, 
TS-VHA-$\mathsf{C1}$-VA4 accumulates more regret and performs poorly relative to TS-VHA-$\mathsf{C1}$-VA$n$, $n=1,2,3$. 

\begin{figure}[hbt]
    \centering
    \includegraphics[width=16.5pc]{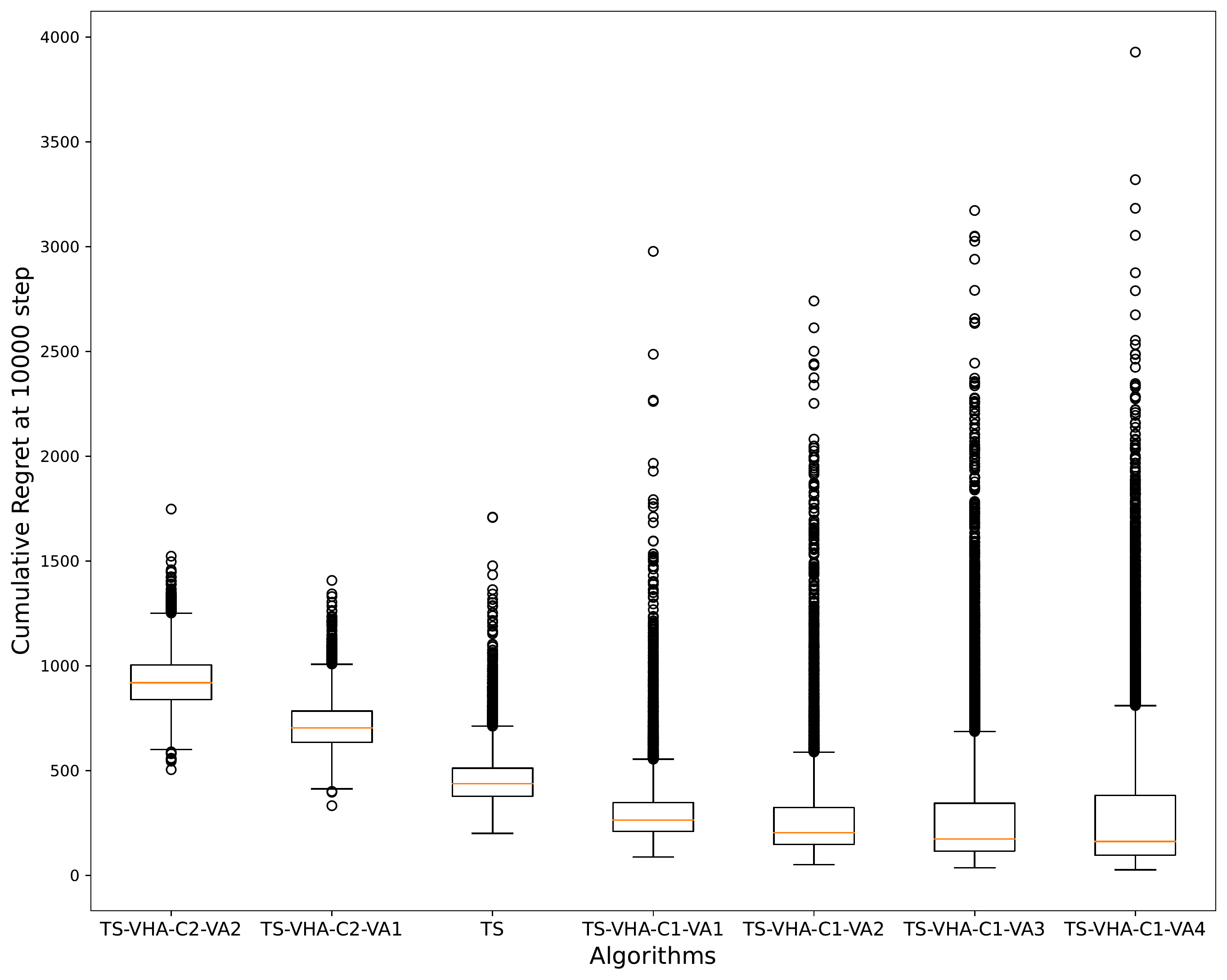}
    \caption{Variation in the cumulative regret with TS,  TS-VHA-$\mathsf{C1}$ and TS-VHA-$\mathsf{C2}$ for Gaussian bandits.}
    \label{fig:RandGbox}
\end{figure}

Fig. {\ref{fig:RandGbox}} shows the distribution of final cumulative regret at the end of 10000 time steps, over 1000 runs, for a Gaussian bandit with 20 arms. Reward from arm $i$ is distributed as $\mathcal{N}(\mu_i,1)$, where $\mu_i, i=1,\ldots,20$, is chosen by sampling independently from $\mathcal{U}[0,1]$ once at the beginning of the experiment and kept constant throughout the 1000 runs. TS-VHA-$\mathsf{C1}$ has a higher variance in its final cumulative regret and is thus not suitable for risk-sensitive scenarios. On the other hand, TS-VHA-$\mathsf{C2}$ results in a lower variance in its final cumulative regret and may be preferred in risk-averse applications.  
\subsection{Bernoulli Bandits}
\begin{figure*}[t]
    \centering
    \subfigure[20 Arms]{%
    \includegraphics[width=16.5pc]{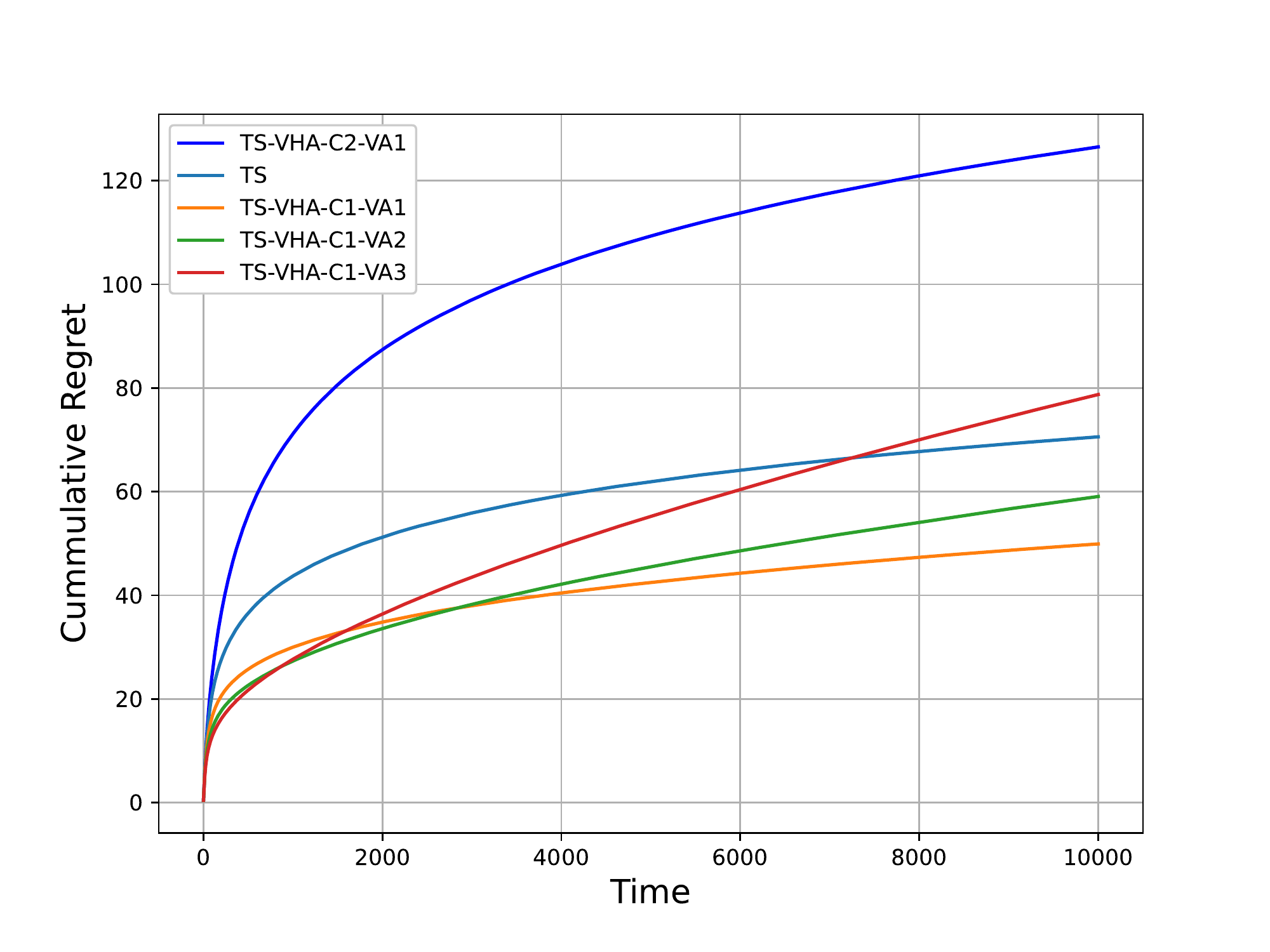}
    \label{fig:RandBsubfig_a}}
    \quad
    \subfigure[200 Arms]{%
    \includegraphics[width=16.5pc]{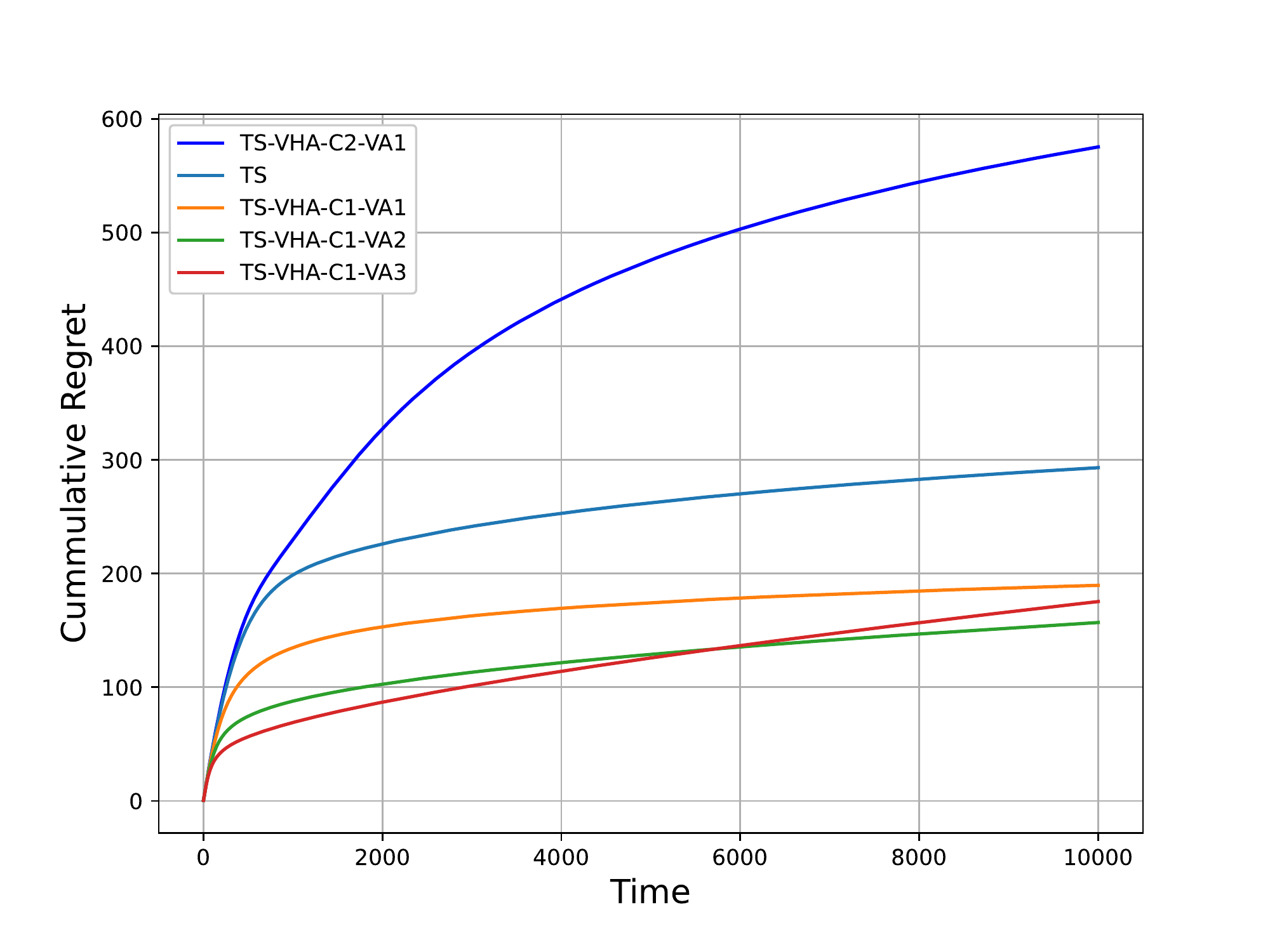}
    \label{fig:RandBsubfig_b}}
    \caption{Bernoulli bandit: Cumulative regret comparison of TS with TS-VHA-$\mathsf{C1}$ and TS-VHA-$\mathsf{C2}$.}
    \label{fig:RandB}
\end{figure*}
We now evaluate the performance of TS-VHA over Bernoulli bandits, i.e., bandit problems with Bernoulli distributed rewards and Beta distribution as the prior.
\subsubsection{Bernoulli Bandit with Randomized Mean Rewards}
Similar to the Gaussian bandits discussed above, we consider two Bernoulli bandit problems, one with 20 arms and the other with 200 arms, with mean reward of each arm is independently sampled from $\mathcal{U}[0,1]$. 

Fig.~\ref{fig:RandBsubfig_a} and Fig.~\ref{fig:RandBsubfig_b} shows the cumulative regret over 100000 time steps, averaged over 1000 independently sampled problem instances, for the 20 armed bandit and the 200 armed bandit, respectively. 

\subsubsection{Real World Datasets}
Here, we show the effectiveness of TS-VHA-$\mathsf{C1}$ on the real-world data sets {\em{Coupon-Purchase}}~\cite{Kaggle2016} and {\em{edX-Courses}}~\cite{Chuang2016}. 

\begin{figure*}[htb]
    \centering
    \subfigure[Considering coupon purchase rate as the mean reward of each arm.]{%
    \includegraphics[width=16.5pc]{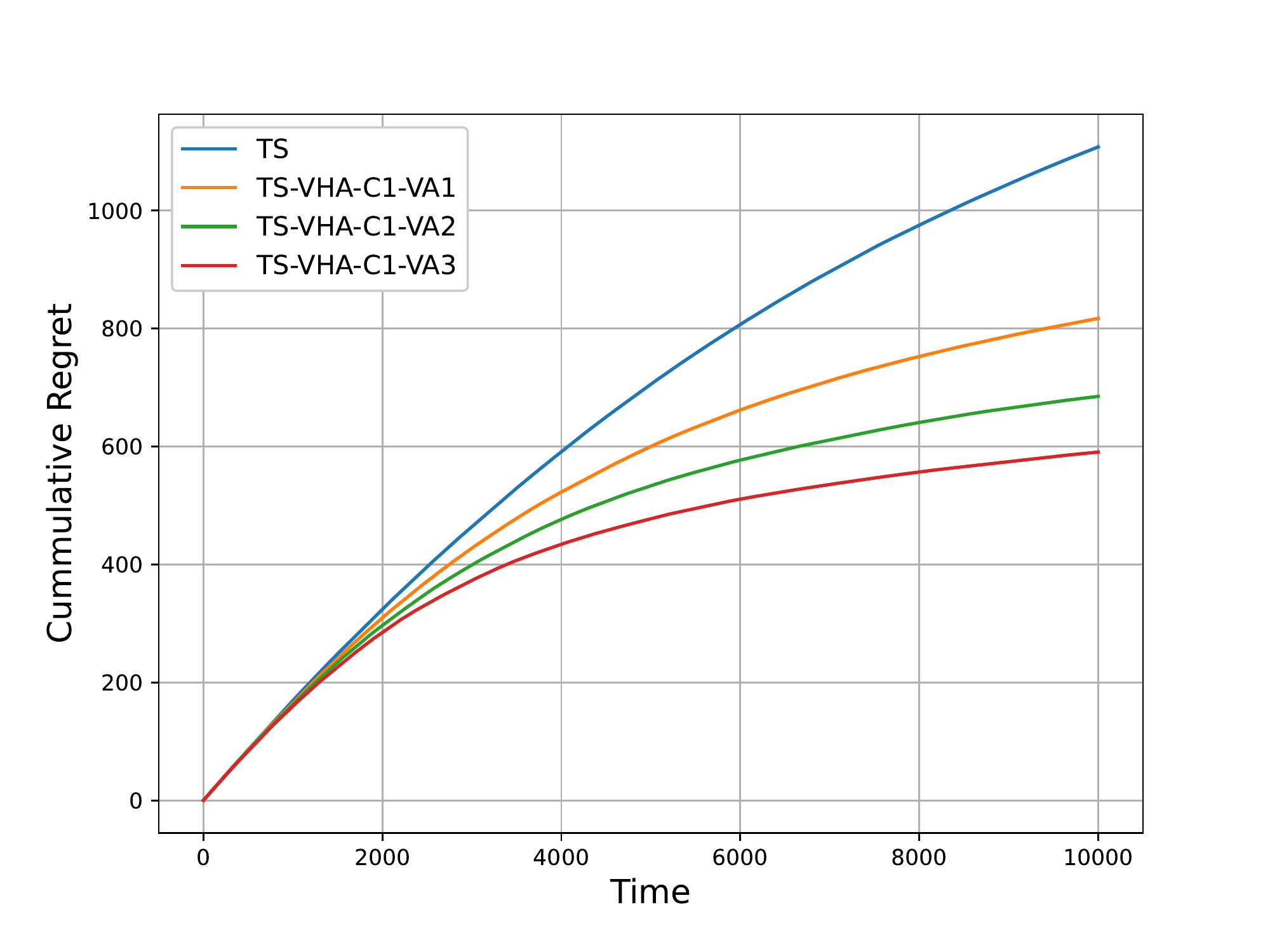}
    \label{fig:coupon_a}}
    \quad
    \subfigure[Considering coupon purchase rate multiplied by the normalized selling price as the mean reward of each arm.]{%
    \includegraphics[width=16.5pc]{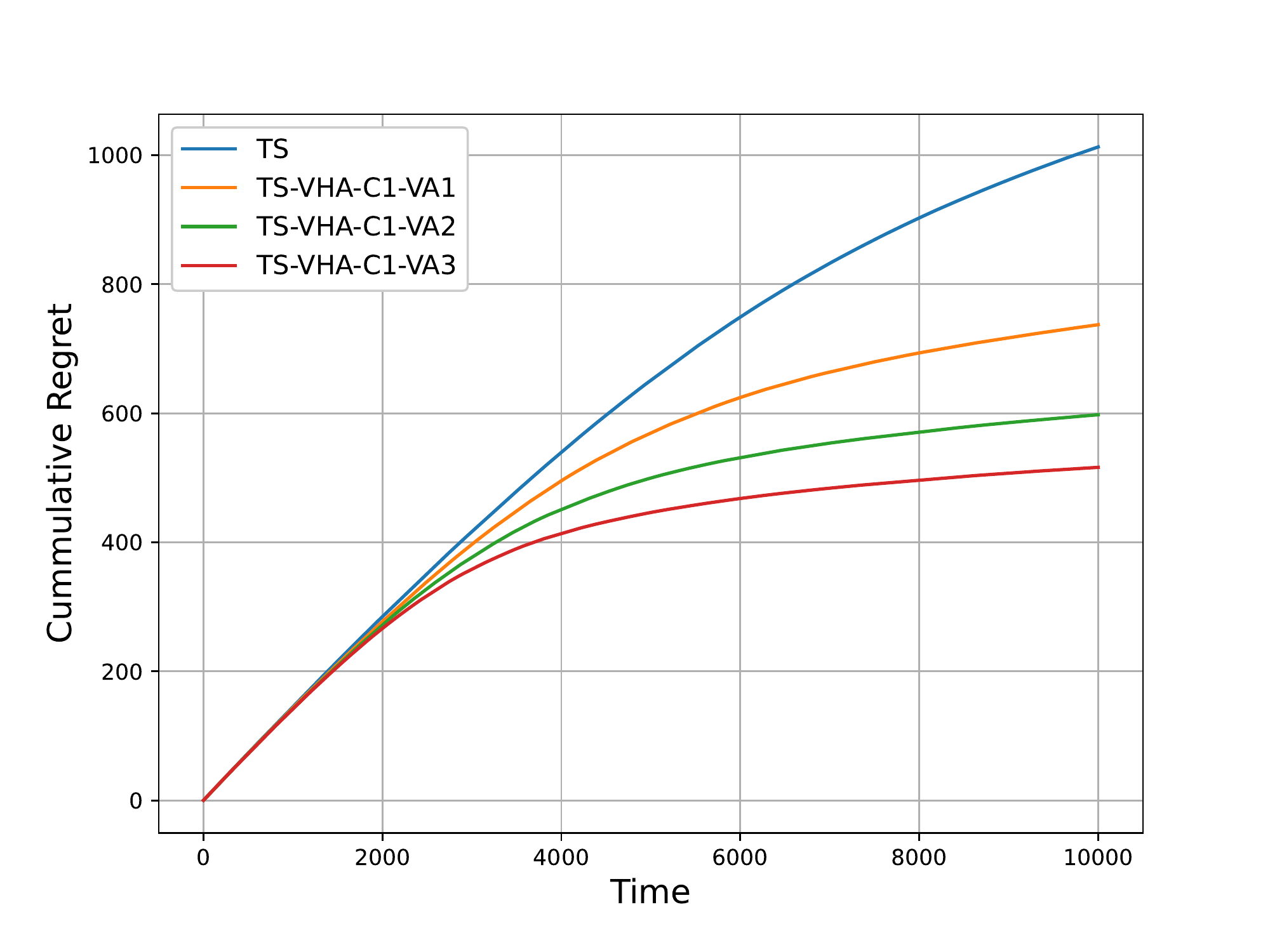}
    \label{fig:coupon_b}}
    \caption{Bernoulli bandit problem with 142 arms, formulated using the Coupon-Purchase dataset.}
    \label{fig:coupon}
\end{figure*}

The Coupon-Purchase dataset contains discount coupons applied to online purchases. From the dataset, we have considered only 142 coupons that correspond to products priced less than or equal to 200 price units {\em{and}} purchased by at least one customer (as in~\cite{Saxena2020}). For these 142 coupons, we have extracted the purchase rate that lies within $[0, 0.3]$ and the final selling price normalized by 200 price units, which lies within $(0, 1]$. 
With each coupon as an independent arm that (when played) generates a binary valued reward according to a Bernoulli distribution, we formulate two bandit problems. In the first one, the mean reward of an arm is equal to the corresponding coupon purchase rate 
and, in the second problem, the mean reward of each arm is equal to the coupon  purchase rate multiplied by the corresponding selling price normalized by 200. By modeling the mean reward of each arm using a Beta distribution, we present the performance of TS and TS-VHA-$\mathsf{C1}$-VA$n$, $n=1,2,3$, in Fig.~\ref{fig:coupon_a} and Fig.~\ref{fig:coupon_b}, corresponding to the first and the second problem, respectively. As can be seen, TS-VHA with combiner $\mathsf{C1}$ helps achieving a lower cumulative regret for both the problems.

\begin{figure*}[!htb]
    \centering
    \subfigure[Considering the course certification rate as the mean reward of each arm.]{%
    \includegraphics[width=16.5pc]{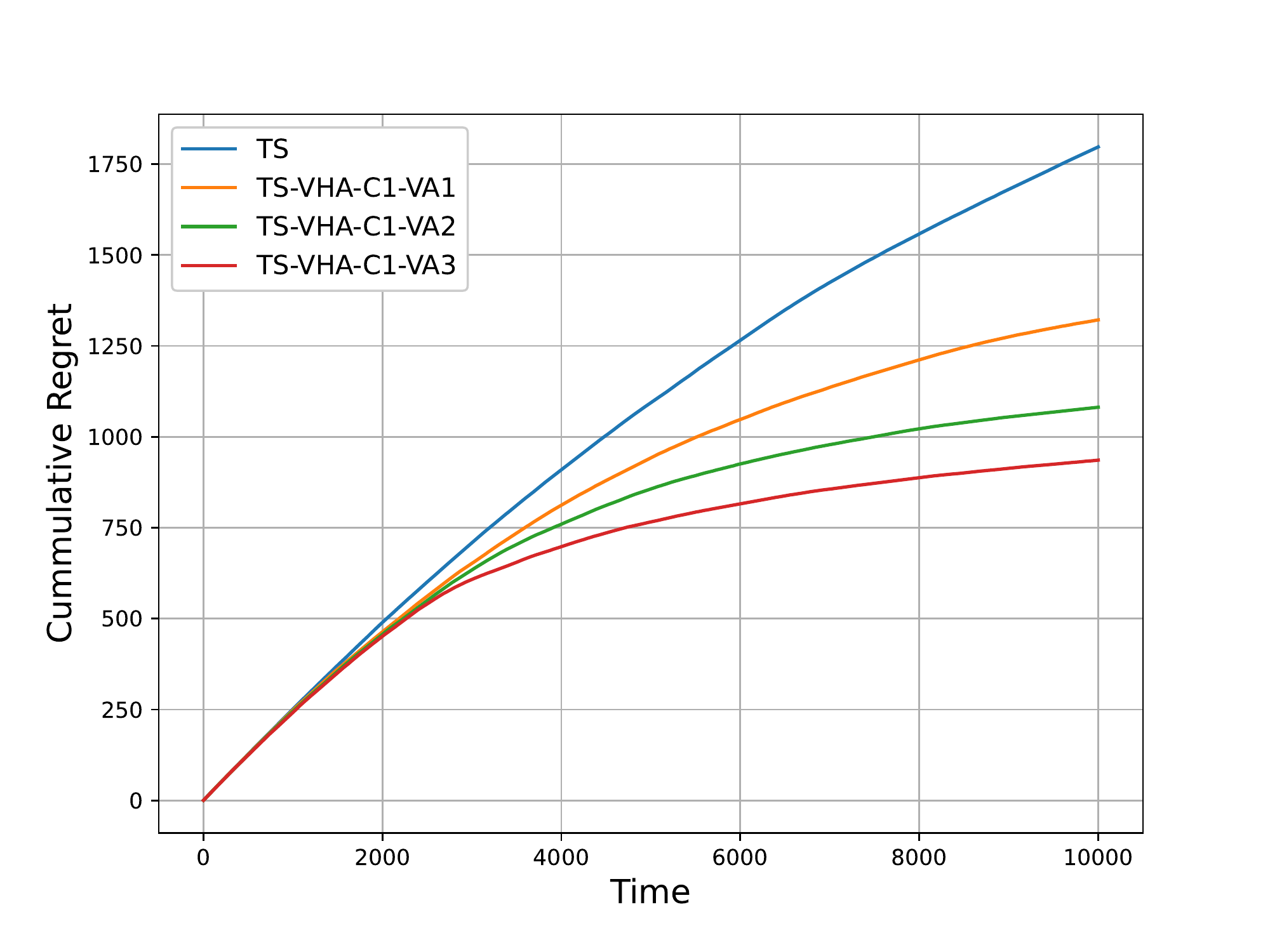}
    \label{fig:edx_a}}
    \quad
    \subfigure[Considering the course certification rate multiplied by the course participation rate as the mean reward of each arm.]{%
    \includegraphics[width=16.5pc]{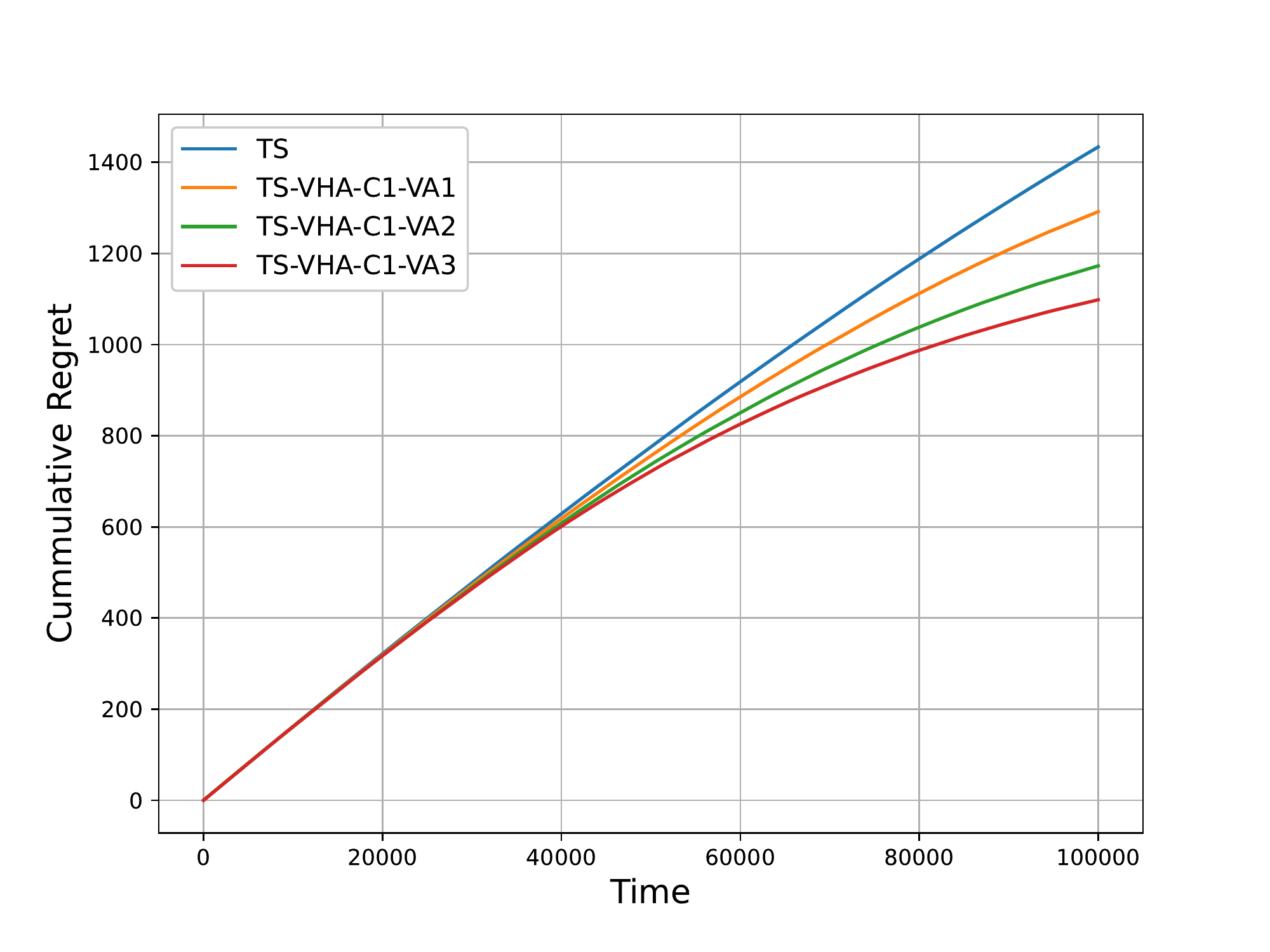}
    \label{fig:edx_b}}
    \caption{Bernoulli bandit problem with 290 arms, formulated using the edX-Course datset.}
    \label{fig:edx}
\end{figure*}

The edX-Courses dataset contains information regarding 290 Harvard and MIT courses and, as in~\cite{Chen2018}, we compute the normalized course participation rates (that lie within unit interval) through min-max normalization of the number of participants in each course and obtain the course certification rates by dividing the number of certified participants in each course by the number of course participants. We formulate two bandit problems by considering each course as an independent arm that returns a Bernoulli distributed reward. In the first problem, the mean reward of each arm is given by the course certification rate and in the second, course certification rate multiplied by the course participation rate is the mean reward. 

With Beta distribution as the prior for the mean reward of each arm, Fig. \ref{fig:edx_a} and Fig. \ref{fig:edx_b} compare the cumulative regret performance of TS and and TS-VHA-$\mathsf{C1}$-VA$n$, $n=1,2,3$, for the first and the second problem, respectively. 

\begin{figure*}[!htb]
    \centering
    \subfigure[Deterministic bandit.]{%
    \includegraphics[width=14.5pc]{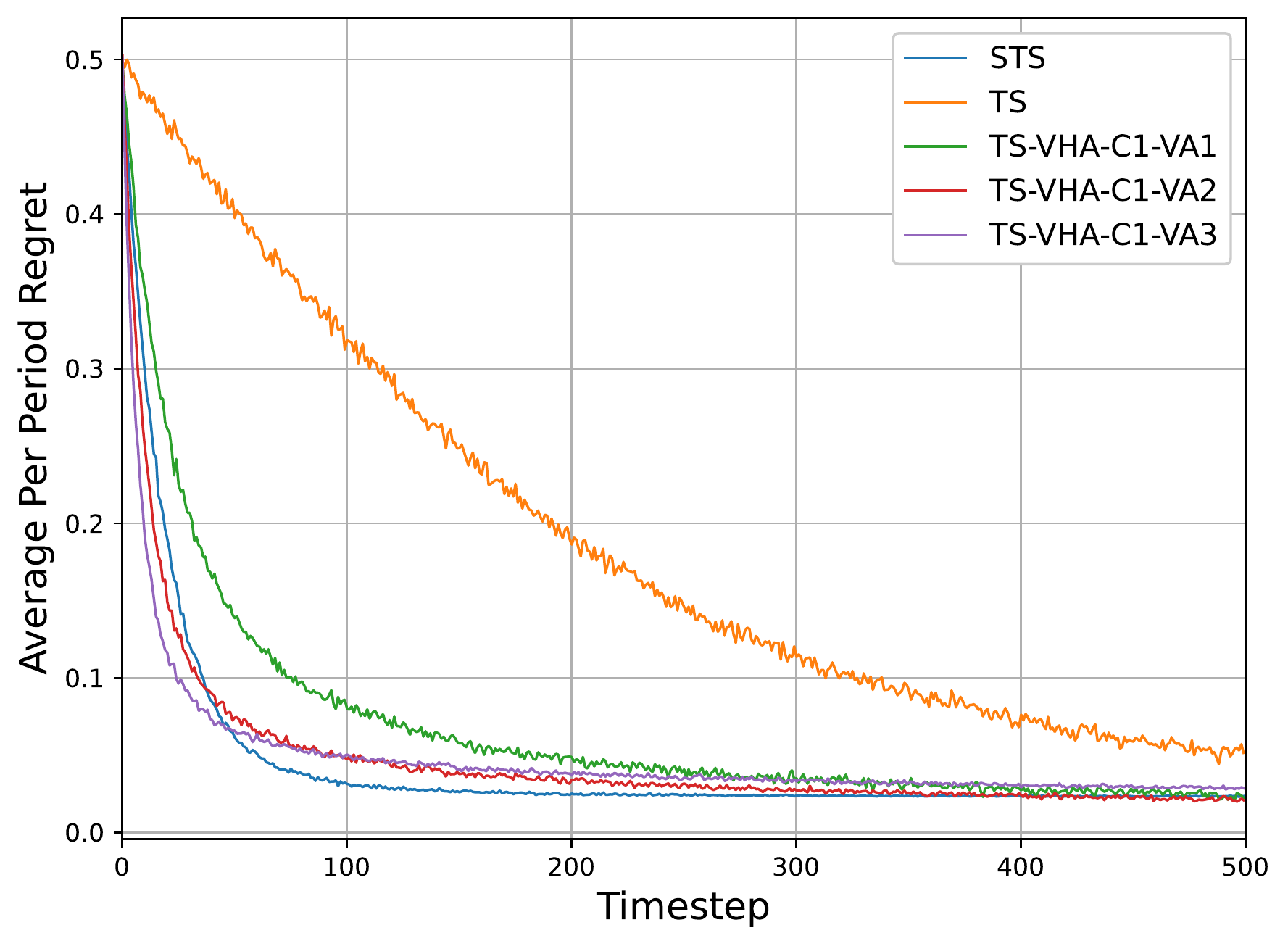} 
    \label{fig:STS_a}}
    \subfigure[Bernoulli bandit]{%
    \includegraphics[width=14.5pc]{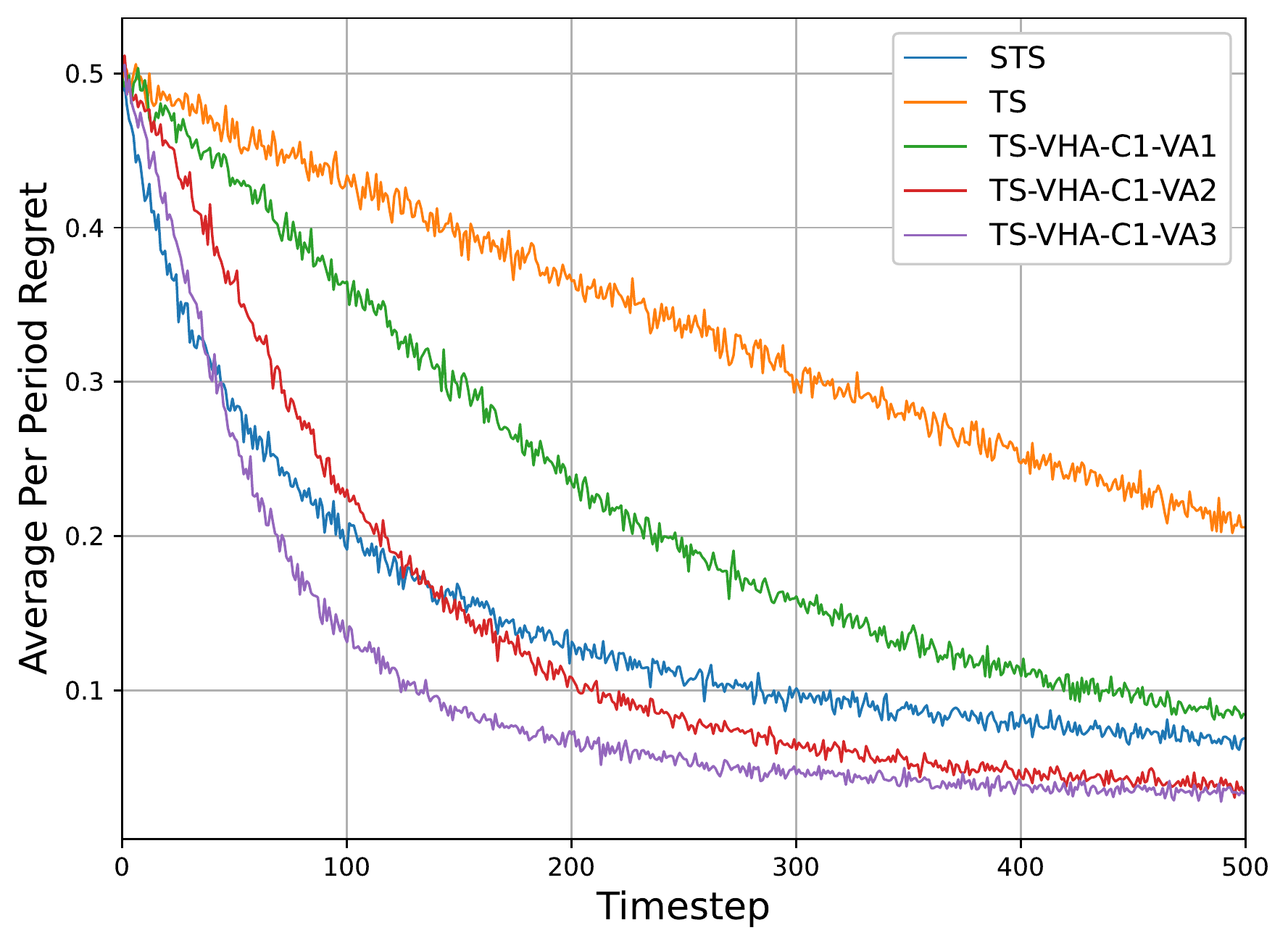} 
    \label{fig:STS_b}}
    \subfigure[Independent Gaussian]{%
    \includegraphics[width=14.5pc]{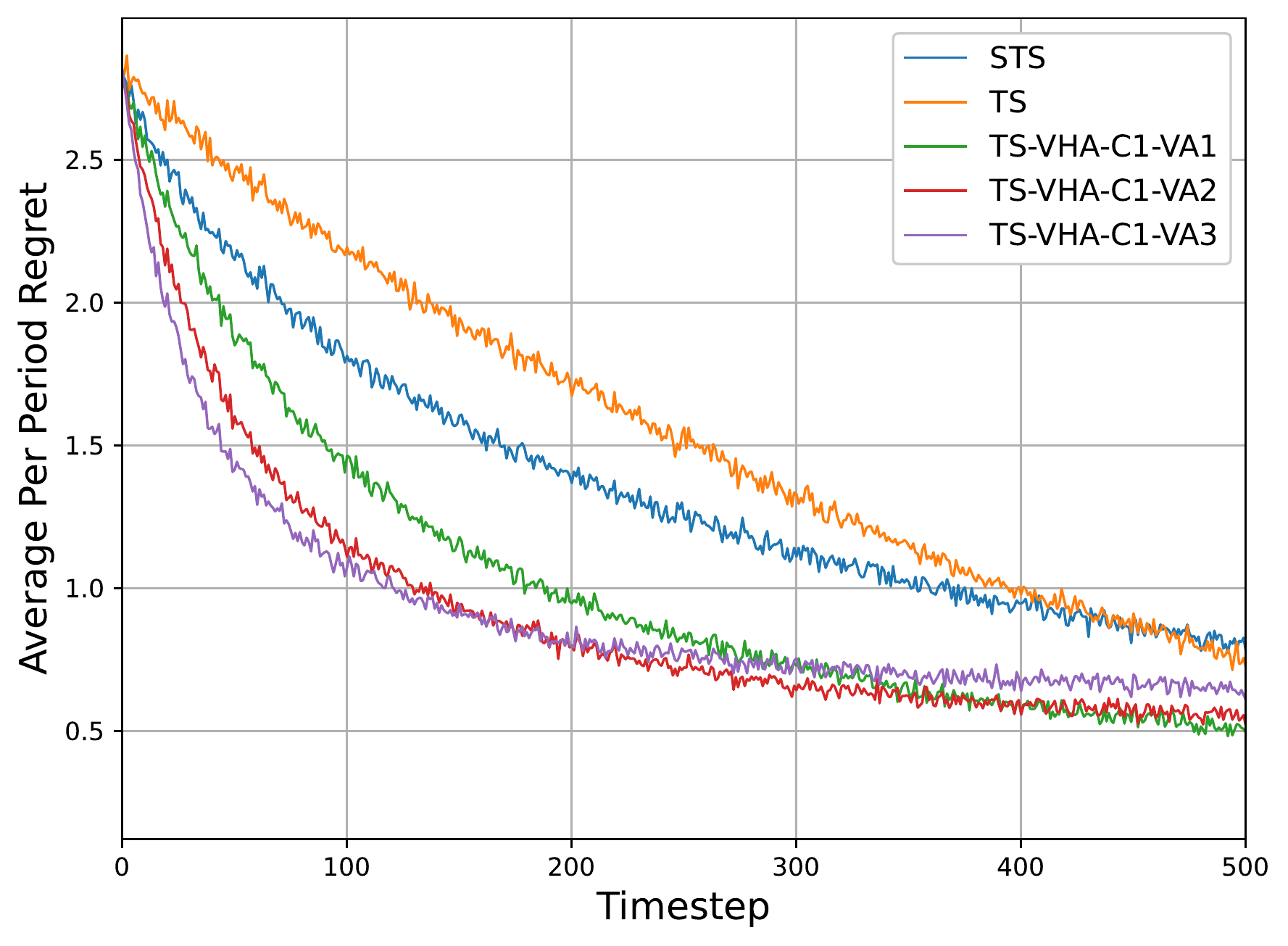}
    \label{fig:STS_c}}
    \subfigure[Linear Gaussian]{%
    \includegraphics[width=14.5pc]{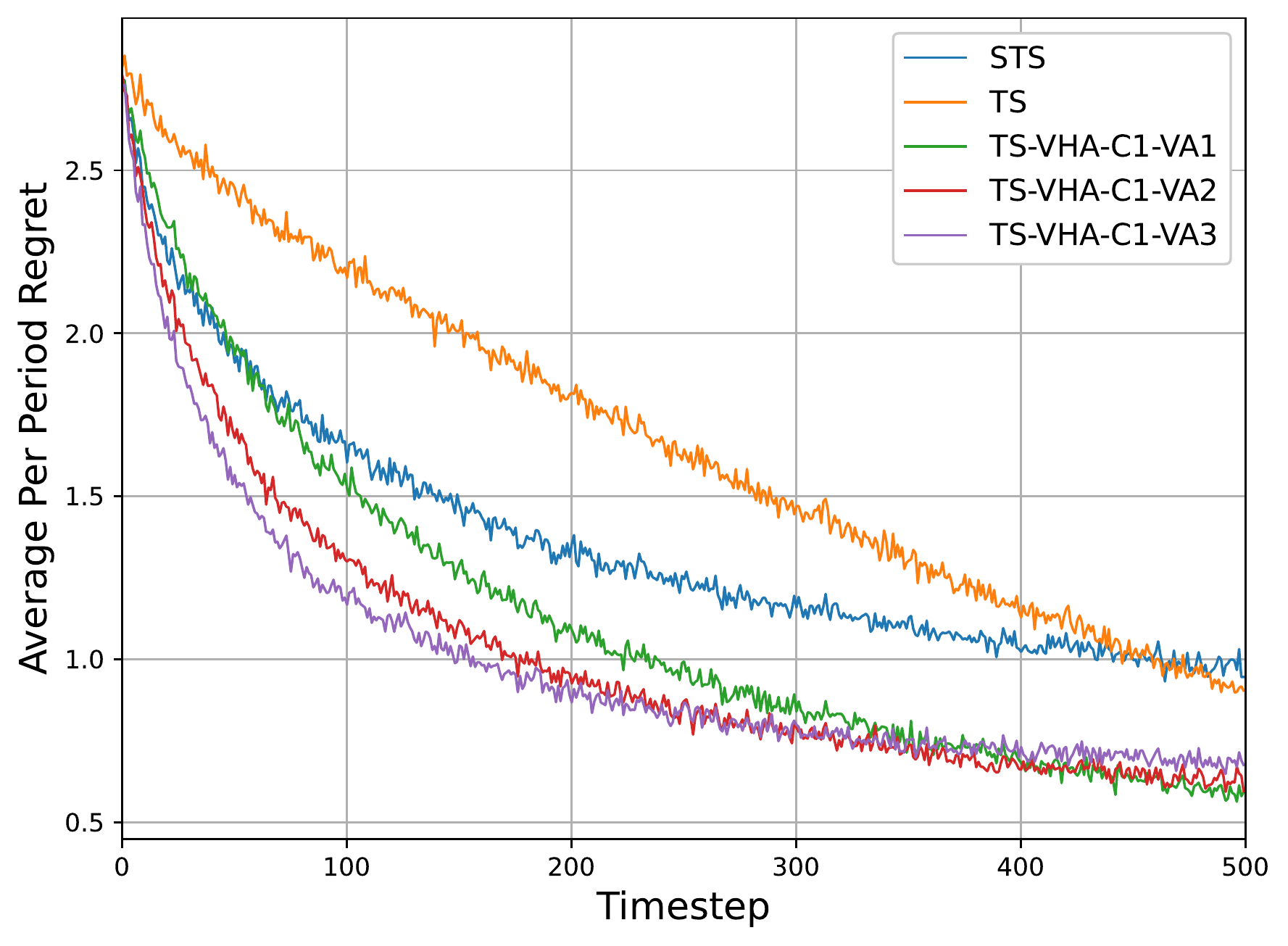}
    \label{fig:STS_d}}
    \caption{Time-sensitive Bandit Learning}
    \label{fig:timesensitive}
\end{figure*}

\begin{figure*}[!htb]
    \centering
    \subfigure[Gaussian bandit with two arms. Mean rewards: 0.5, 0.25.]{\includegraphics[width=16.5pc]{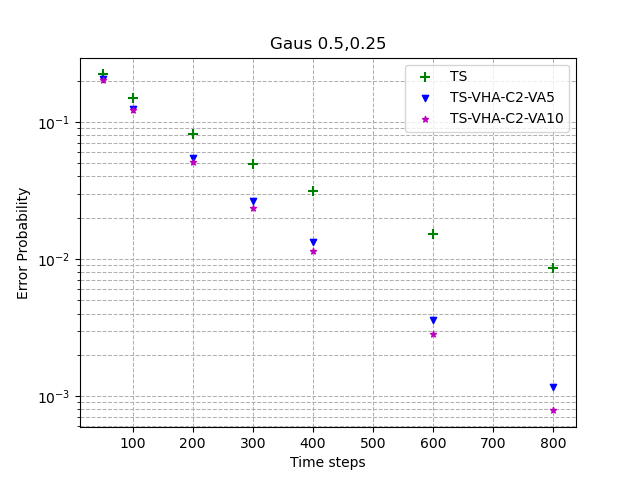}}
    \subfigure[Bernoulli bandit with two arms. Mean rewards: 0.51,  0.5.]{\includegraphics[width=16.5pc]{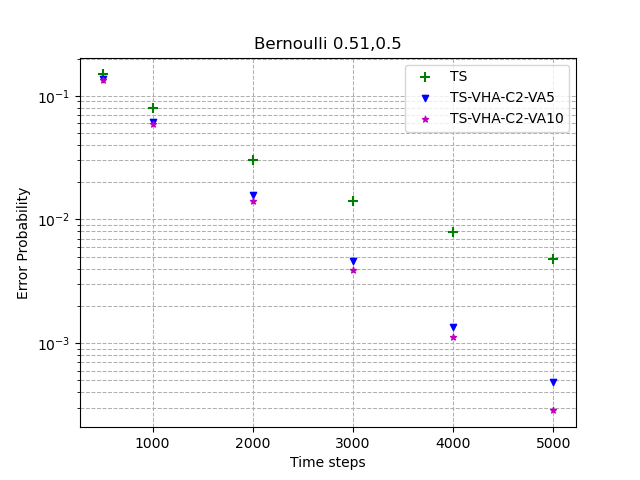}} 
    \caption{Best arm identification for two armed Bernoulli and Gaussian bandits.}
    \label{fig:BestArm}
\end{figure*}

\begin{figure*}[!htbp]
    \centering
    \subfigure[Bernoulli bandit, 2 arms]{%
    \includegraphics[width=14.5pc]{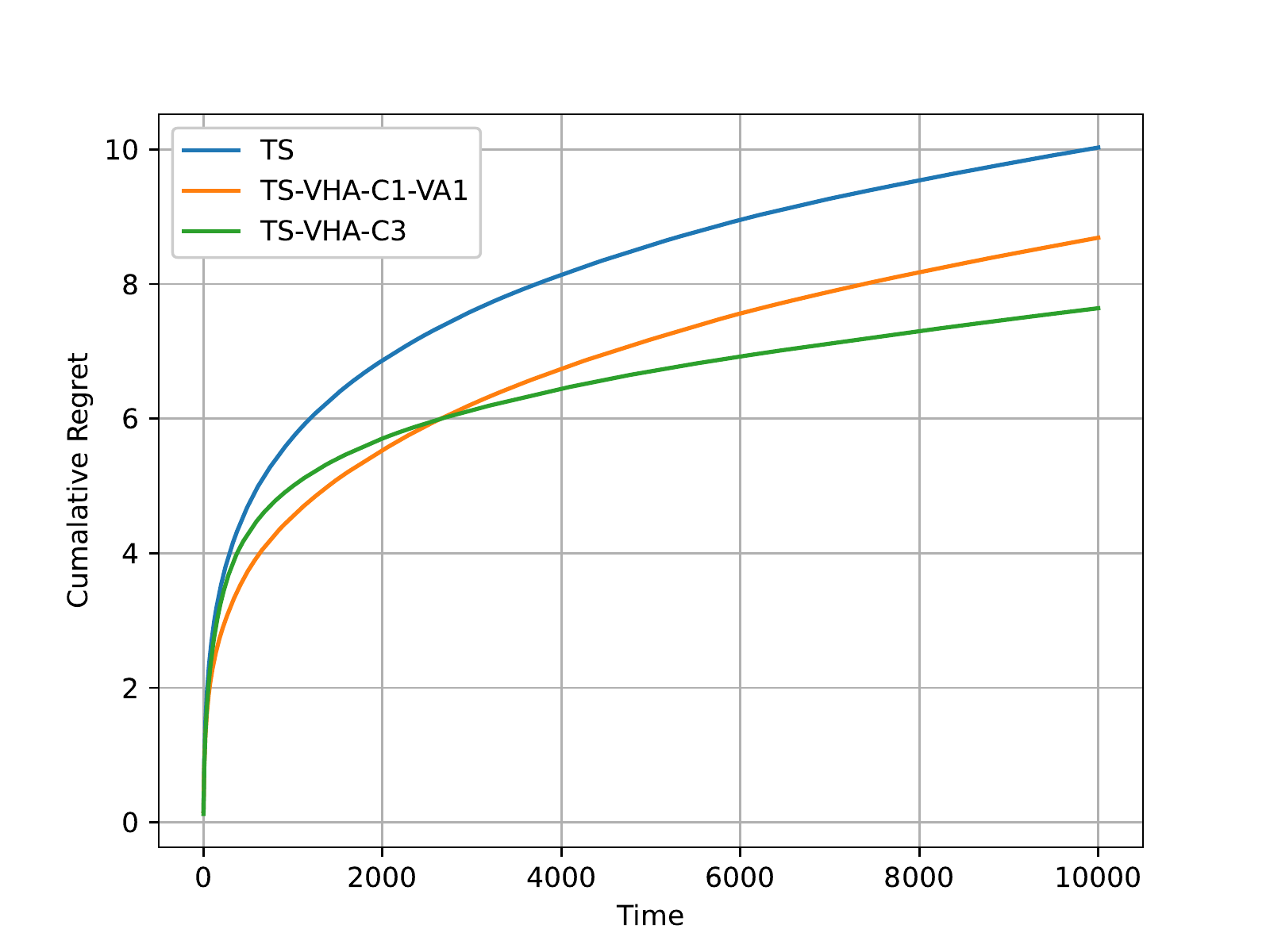} 
    \label{fig:C3_a}}
    \subfigure[Bernoulli bandit, 20 arms]{%
    \includegraphics[width=14.5pc]{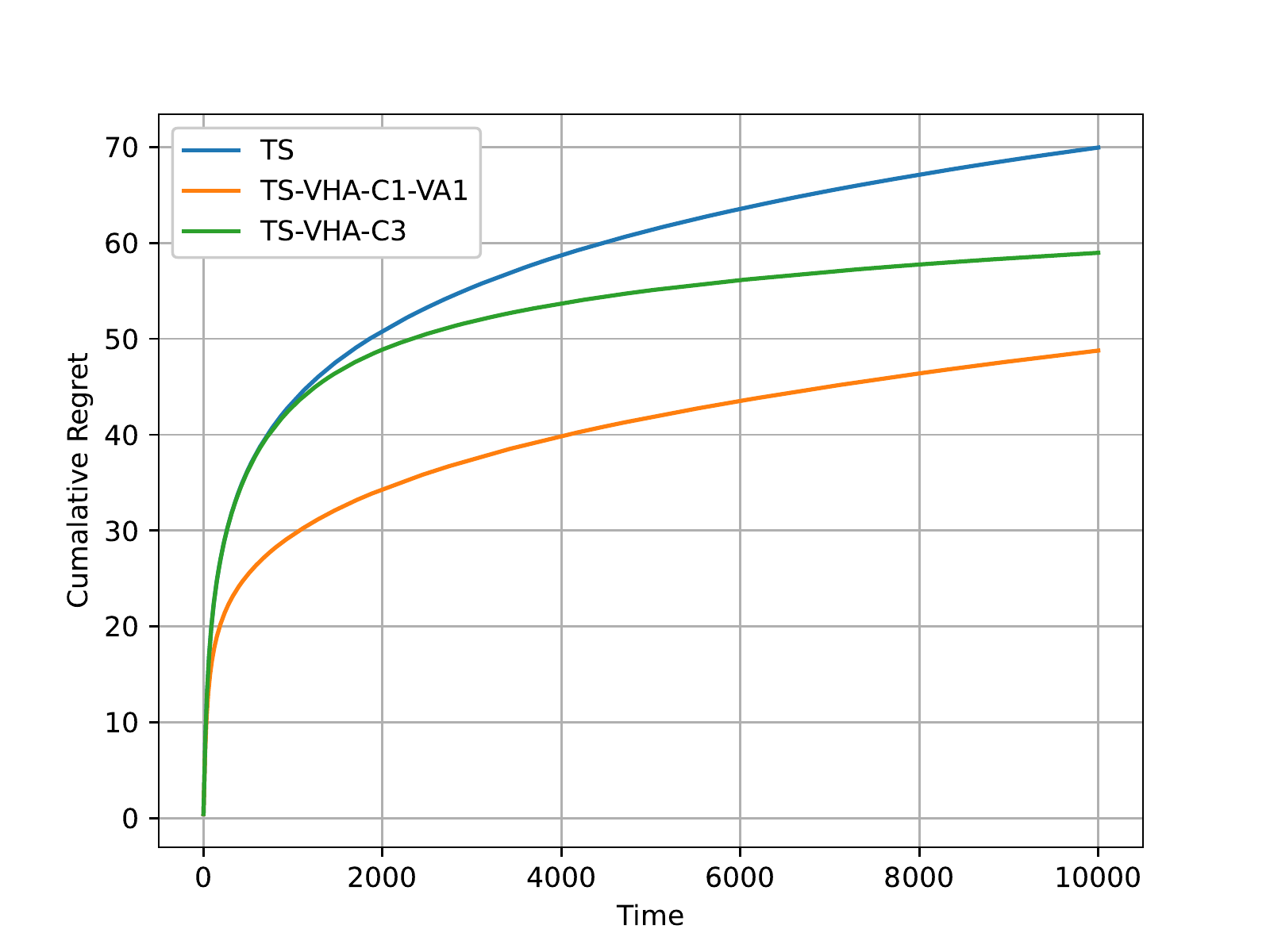}
    \label{fig:C3_b}}
    \subfigure[Gaussian bandit, 2 arms]{%
    \includegraphics[width=14.5pc]{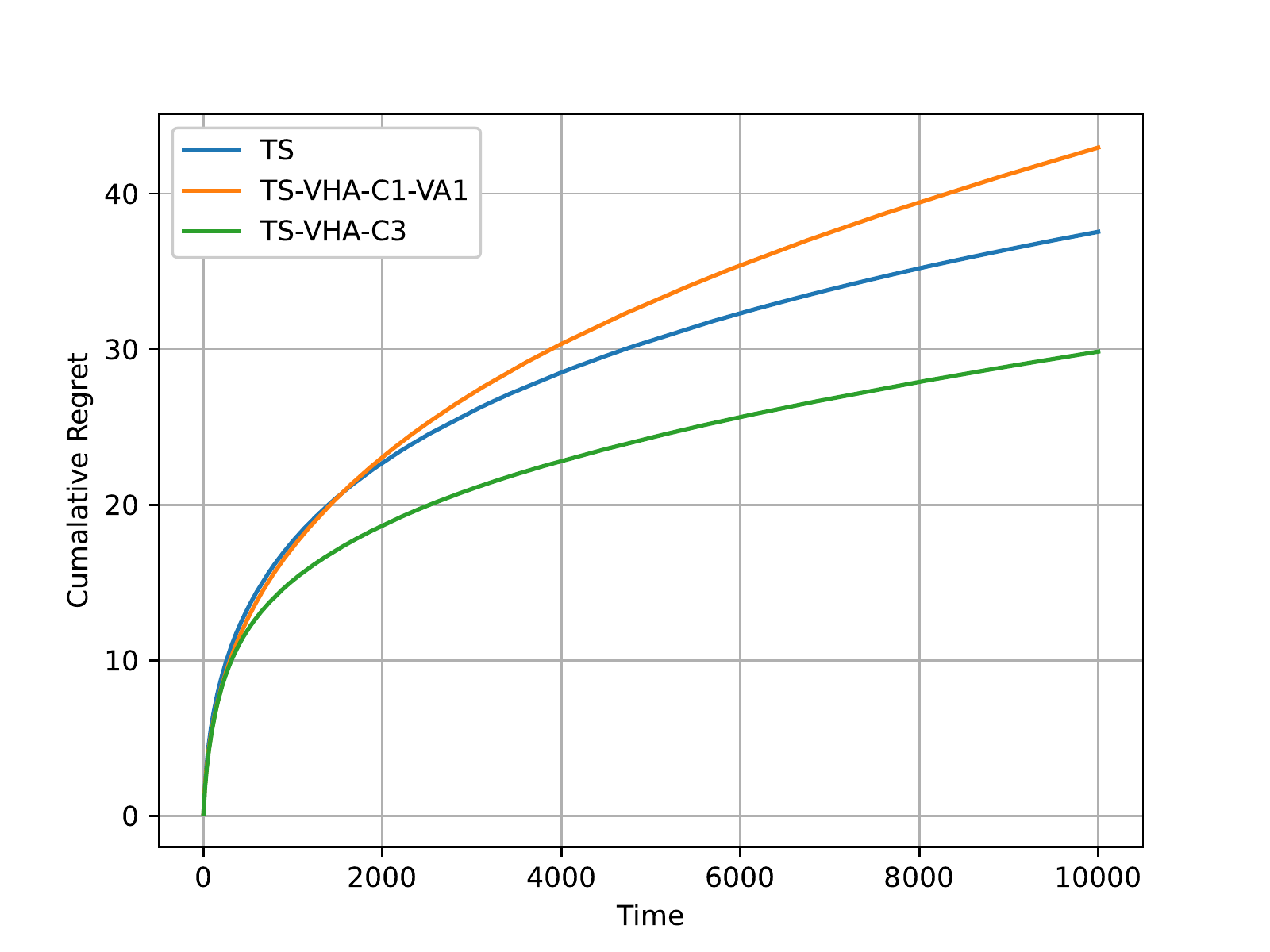}
    \label{fig:C3_c}}
    \subfigure[Gaussian bandit, 20 arms]{%
    \includegraphics[width=14.5pc]{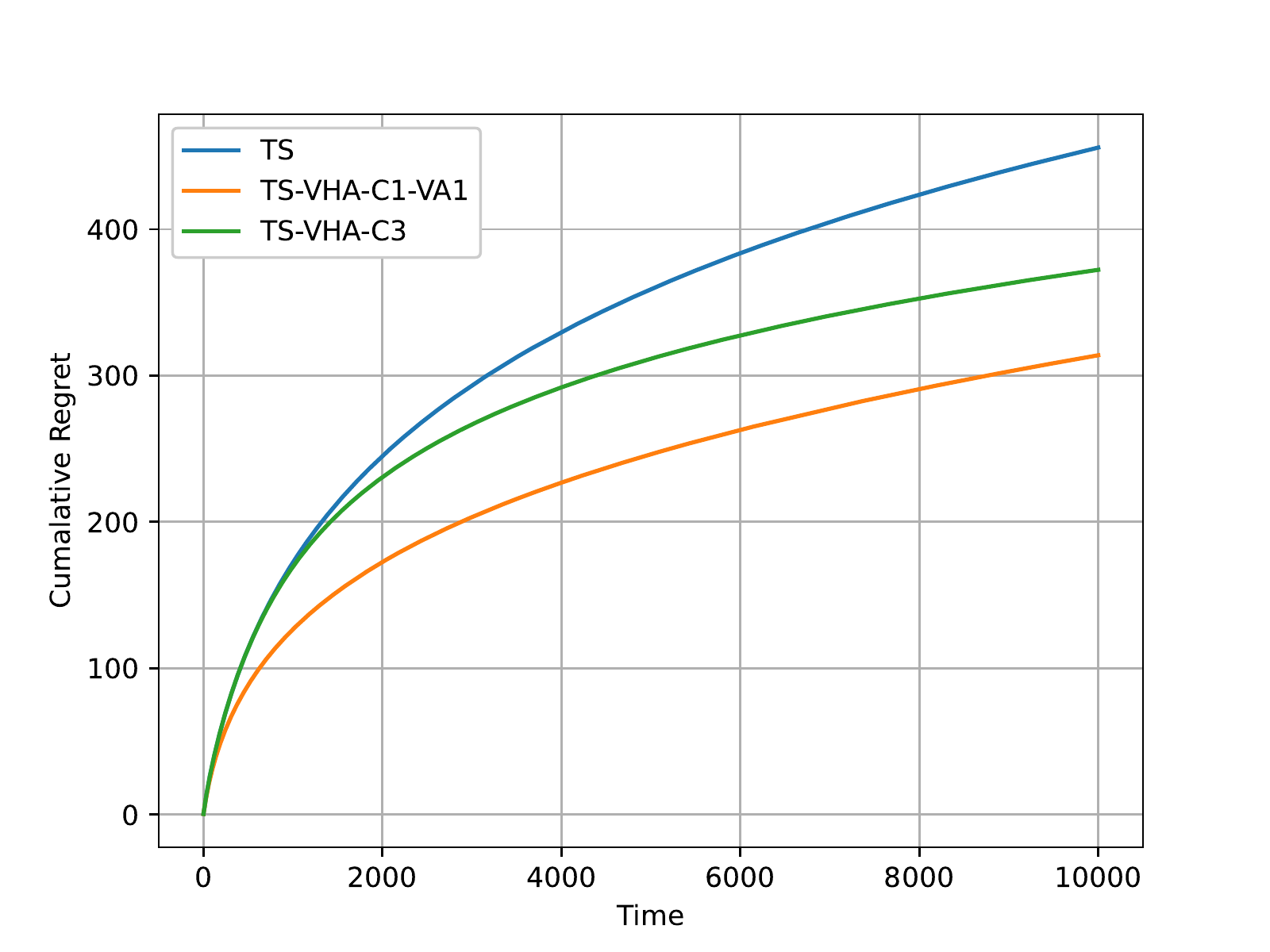}
    \label{fig:C3_d}}
    \caption{Cumulative regret comparison of TS-VHA-$\mathsf{C3}$ with TS and TS-VHA-$\mathsf{C3}$}
    \label{fig:timesensitive}
\end{figure*}

\subsection{Time-Sensitive Bandit Learning}

Most of the bandit algorithms focus on learning the optimal arm (or, action). Often, especially for bandit problems having a very large set of arms, convergence to optimality may take a long time rendering them not useful in some  practical applications. 
For example, in the case of a recommender system, the learning agent may be required to impress upon the users through its near optimal recommendations during the early interactions; Or, the learning agent may not have enough number of interactions with each user to converge onto perfect recommendations.

In~\cite{Russo2020}, the authors have addressed the problem of learning near-optimal {\em{satisficing}} actions 
considering situations where the near term performance is more important than the performance over an asymptotically long time horizon, or, the optimal action is costly to learn relative to near-optimal actions. Satisficing Thompson Sampling (STS), proposed in~\cite{Russo2020}, performs {\em{time-sensitive}} learning by modifying the {\em{Select action}} step of TS. 

Recall that, in TS (i.e., Algorithm~\ref{alg:TS}), $\theta_i(\tau)$ is the sample drawn from posterior of arm $i$ at time $\tau$, $i(\tau)$ is the index of arm played at time $\tau$ and $\theta_{i(\tau)}$ is the expected reward of the arm played at time $\tau$. At each time step $t$, STS identifies an $\epsilon$-optimal action through the following {\em{Select action}} step.

\noindent{\bf{Select action (in STS):}}
\begin{itemize}
    \item[] Let $i(t) = \arg \max_i \theta_i(t)$.
    \item[] Let $ {\hat{\tau}} = \min \{ \tau \in \{1, \ldots, t-1\}: \theta_{i(\tau)} + \epsilon \geq \theta_{i(t)}\}$.
    \item[] If $\hat{\tau}$ is not null, then $i(t)=i(\hat{\tau})$.
\end{itemize}
Essentially, at each time step $t$, STS chooses to play an arm $k$ that has already been played in the past, as long as the estimate of the expected reward from arm $k$ is not lower than the estimate of the expected reward from an optimal arm (optimal at time $t$ as per the TS) by $\epsilon$ units. Thus, STS exploits more by re-using near-optimal satisficing arms rather than exploring un-used arms. 
With {\em{per period regret}} as the performance metric (that captures the {\em{time preference}}), the simulation experiments reported in~\cite{Russo2020} show that STS can significantly outperform TS when the optimal action is costly to learn relative to satisficing near-optimal actions. 
We consider four simulation experiments that are same as those considered in~\cite{Russo2020} and compare the performance of TS-VHA-$\mathsf{C1}$ with that of STS and TS in Fig.~\ref{fig:timesensitive}. For all the four experiments, we compute the per period regret over 500 time steps, averaged over 5000 runs. Fig.~\ref{fig:STS_a} considers a deterministic bandit with 250 arms with mean reward of each arm sampled independently from $\mathcal{U}[0,1]$. As every arm, when played, returns the reward equal to its mean reward, it is referred to as a deterministic bandit. Performance of TS-VHA-$\mathsf{C1}$-VA2 and TS-VHA-$\mathsf{C}1$-VA3 is very close to that of STS. Fig.~\ref{fig:STS_b} corresponds to a bandit that differs from that of Fig.~\ref{fig:STS_a} as follows: Whenever an arm is played, the observed reward is a Bernoulli random variable with success probability equal to the mean reward. Note that, in Fig.~\ref{fig:STS_a} and Fig.~\ref{fig:STS_b}, we consider $\epsilon=0.05$ for the STS. It can be observed that TS-VHA-$\mathsf{C1}$-VA2 and TS-VHA-$\mathsf{C1}$-VA3 perform better than STS for time step values above (approximately) 150 and 50, respectively. Fig.~\ref{fig:STS_c} corresponds to a 250 armed Gaussian bandit with mean reward of each arm sampled independently from $\mathcal{N}(0,1)$; When an arm is played, the realized reward is the sum of the arm's mean reward and an independent sample from $\mathcal{N}(0,1)$. Here, $\epsilon=0.5$ for the STS. 
Finally, we consider linear Gaussian bandit with 250 arms in Fig.~\ref{fig:STS_d}. The mean rewards are given by the vector $\mathbf{L}\boldsymbol{\theta}\in\mathbb{R}^{250\times 1}$, where $\boldsymbol{\theta}\in \mathbb{R}^{250\times 1}$ is sampled from $\mathcal{N}(0,\mathbf{I})$ and $\mathbf{L}\in \mathbb{R}^{250\times 250}$ is a random matrix with each row drawn independently and uniformly from the unit sphere. While $\boldsymbol{\theta}$ is unknown a priori, $\mathbf{L}$ is known before hand. When an arm is played the observed reward is the sum of the mean reward and an independent sample from $\mathcal{N}(0,2)$. 
As can be observed from Fig.~\ref{fig:STS_c} and Fig.~\ref{fig:STS_d}, TS-VHA-$\mathsf{C1}$ outperforms TS and STS for both independent Gaussian and Linear Gaussian bandits. 

We think, the above simulation experiments {\em{only}} indicate that it might be interesting to investigate (and analyze) the TS-VHA-$\mathsf{C1}$ from the aspect of time-sensitive learning. 

\subsection{Best Arm Identification}
Next, we consider the {\em{fixed budget setting}} of the {\em{Best Arm Identification}} problem as discussed in~\cite{BestArmKauffman}. The idea is to identify the best arm amongst all the bandit arms by playing them intelligently for a fixed number of time steps $t$. The metric used to compare algorithms is the probability of error in identifying the best arm after the fixed time step $t$.

TS can be utilized to solve this problem by designating the arm with the highest empirical mean after $t$ time steps as the best arm. However, TS performs poorly for this pure-exploration problem because of its high exploitative nature. Therefore, with the intention to increase the exploration in TS, we evaluate the applicability of TS-VHA-$\mathsf{C2}$ in this scenario.

In Fig.~\ref{fig:BestArm}, we consider the Bernoulli bandit as well as Gaussian bandit, each having two arms. The plots on the left and right have arms with mean rewards equal to $(0.5, 0.25)$ and $(0.51, 0.5)$, respectively. For both the scenarios, we observe that TS-VHA-$\mathsf{C2}$ outperforms TS empirically. 

\subsection{Combiner $\mathsf{C3}$}
Finally, we evaluate the cumulative regret performance of Combiner $\mathsf{C3}$, through simulations, for Bernoulli bandits and Gaussian bandits and compare its performance with TS and TS-VHA-$\mathsf{C1}$.

Similar to section \ref{sec:sims}-A and \ref{sec:sims}-B, we first evaluate the performance of $\mathsf{C3}$ on the randomized 20 arms case. As shown in Fig. \ref{fig:C3_b} and Fig. \ref{fig:C3_d}, for both Gaussian and Bernoulli bandits, TS-VHA-$\mathsf{C3}$ outperforms TS. 

Next, we choose the same randomized scenario but with two arms in Fig. \ref{fig:C3_a} and Fig. \ref{fig:C3_c}. In this case, $\mathsf{C3}$ outperforms both TS and TS-VHA-$\mathsf{C1}$ significantly. Interestingly, for the Gaussian bandits, TS-VHA-$\mathsf{C1}$ performs inferior to TS, suggesting that, in some cases, increasing exploitation from the beginning does not help in optimizing the cumulative regret. But, dynamically adjusting the amount of exploitation over time by $\mathsf{C3}$ provides superior performance. A mathematical analysis of the regret bound for $\mathsf{C3}$ would help gaining more insight into it.  

\section{Conclusion}
\label{sec:conclusion}
We have proposed a general framework, Thompson Sampling with Virtual Helping Agents (TS-VHA), that \textit{combines} samples drawn by the \textit{virtual agents} to maneuver the exploration vs exploitation tradeoff in Thompson Sampling. Based on this framework, we developed two linear combiners (TS-VHA-$\mathsf{C1}$ and TS-VHA-$\mathsf{C2}$) and analysed theoretically their cumulative regret performance on Gaussian Bandits. Moreover, we showed their empirical efficacy on both Gaussian and Bernoulli bandits for multiple metrics: cumulative regret, best-arm identification and time-sensitive learning. We defer the analysis of the regret bounds on these metrics for our future work. To exhibit the broad scope of the framework, we also put forth a nonlinear combiner TS-VHA-$\mathsf{C3}$ that dynamically tunes the amount of exploration/exploitation and offers superior empirical performance. It would be interesting to experiment and devise more sophisticated combiners. TS-VHA can be applied wherever Thompson Sampling can be applied and we leave extending TS-VHA (along with designing combiners) for contextual bandits, non-stationary bandits and restless bandits for future work. Finally, exploring the usage of neural networks in developing combiners would be an exciting avenue for future work.

\appendices
\section{Inequalities used in the regret analysis}
\label{app:inequalities}
\begin{inequality}\label{ineq1} 
(Chernoff-Hoeffding Bound)
Let $X_1, \ldots, X_n$ be independent 0 - 1 r.v.s with $E[X_i]= p_i$ (not necessarily equal). Let $X = \frac{1}{n}\sum_i X_i$, $\mu = E[X]=\frac{1}{n}\sum_{i=1}^{n}p_i$. Then, for any $0<\lambda<1-\mu$,
\begin{equation*}
    \text{Pr}(X \geq \mu + \lambda) \leq e^{-nd(\mu+\lambda,\mu)},
\end{equation*}
and for any $0<\lambda<\mu$,
\begin{equation*}
    \text{Pr}(X \geq \mu - \lambda) \leq e^{-nd(\mu-\lambda,\mu)},
\end{equation*}
where $d(a,b) = a\ln{\frac{a}{b}+(1-a)\ln{\frac{1-a}{1-b}}}$
 
\end{inequality} 

\begin{inequality}\label{ineq2}
(Chernoff-Hoeffding Bound)
Let $X_1, \ldots, X_n$ be random variables with common range $[0,1]$ and such that $\mathbb{E}[X_t | X_1, \ldots, X_{t-1}]=\mu$. Let $S_n = \sum_{i=1}^{n} X_i$. Then, for all $a \geq 0$,
\begin{equation*}
    \text{Pr}(S_n \geq n\mu + a) \leq e^{-2a^2/n},
\end{equation*}
and 
\begin{equation*}
    \text{Pr}(S_n \leq n\mu - a) \leq e^{-2a^2/n}.
\end{equation*}
\end{inequality} 

The following inequalities can be derived for a Gaussian random variable from Formula $7.1.13$ in \cite{Abramowitz}. 

\begin{inequality}\label{ineq3}
For a Gaussian distributed random variable $Z$ with mean $m$ and variance $\sigma^2$,
\begin{equation*}
    \text{Pr}(Z > m+x\sigma) \geq \frac{x}{\sqrt{2\pi} (x^2+1)}e^{-x^2/2}.
\end{equation*}
\end{inequality} 

\begin{inequality}\label{ineq4}
For a Gaussian distributed random variable $Z$ with mean $m$ and variance $\sigma^2$, for any $z$, 
\begin{equation*}
    \frac{1}{4\sqrt{\pi}} e^{-7z^2/2} < Pr (|Z-m| > z\sigma) \leq \frac{1}{2} e^{-z^2/2}.
\end{equation*}
\end{inequality}

\begin{inequality}\label{ineq5}
Let $S_n = \sum_{i=1}^{n} \frac{1}{i^p}$. Then for $0 < p < 1$ from \cite{Pseries},  
\begin{equation*}
    S_n < 1+\frac{(n+1)^{1-p}-1}{1-p}
\end{equation*}
\end{inequality} 

\section{Proof of Lemma 1}
\label{app:lemma1}
Recall that $\theta_i(t)\sim \mathcal{N}\left(\hat{\mu}_{i}(t), \frac{1}{\gamma(k_i(t)+1)}\right)$,  $x_i=\mu_i+\frac{\Delta_i}{3}$,  $y_i=\mu_1-\frac{\Delta_i}{3}$ and $L_i(T) = \frac{2 \ln{T\Delta_i^2}}{\gamma(y_i-x_i)^2}$. Given $\mathcal{F}_{\tau_{k}}$, let $\Theta_{k}$ denote a Gaussian random variable distributed as $\mathcal{N}\left(\hat{\mu}_{1}(\tau_{k}+1), \frac{1}{\gamma(k+1)}\right)$. For convenience, we denote $\hat{\mu}_1(\tau_{k}+1)$ with $\hat{\mu}_{1}$ in the following. 
Let $G_k$ be the geometric random variable representing the number of consecutive independent trials until a sample of $\Theta_{k}$ becomes greater than $y_{i}$. Using $\Theta_k$ and \hyperref[defpit]{Definition 6}, we can write $p_{i,\tau_k+1}=\text{Pr}(\Theta_{k}>y_{i}|\mathcal{F}_{\tau_k})$, and 
\begin{equation*}
\eb\left[\frac{(1-p_{i,\tau_{k}+1})}{p_{i,\tau_{k}+1}}\right] = \eb[\eb[G_k|\ \mathcal{F}_{\tau_{k}}]] = \eb[G_k].   
\end{equation*}
Therefore,
\begin{equation}
    \sum_{k=0}^{T-1}\eb\left[\frac{(1-p_{i,\tau_{k}+1})}{p_{i,\tau_{k}+1}}\right] 
    = \underbrace{\sum_{k=0}^{4L_{i}(T)-1} \eb[G_k]}_{\text{Sum (\ref{eq:sums}a)}}+\underbrace{\sum_{k=4L_{i}(T)}^{T-1} \eb[G_k]}_{\text{Sum (\ref{eq:sums}b)}} \label{eq:sums}
\end{equation}
We will now bound Sum (\ref{eq:sums}a), first term on the RHS of \eqref{eq:sums}.
Let $z=\sqrt{\ln{r^{\beta}}}$, where $r\geq1$ is an integer, $\beta \in [1,2)$, and let $M_{r}$ denote the maximum of $r$ independent samples of $\Theta_{k}$. 
\begin{align}
    \text{Pr}(G_k < r) & \geq \text{Pr}(M_{r} > y_{i}) \notag\\
       & \geq  \text{Pr}\left(M_{r} > \hat{\mu}_{1} + \frac{z}{\sqrt{\gamma(k+1)}} > y_{i}\right)\notag \\
       & = \eb\left[\eb\left[M_{r} >\eta > y_{i} \mid \mathcal{F}_{\tau_{k}}\right]\right]\notag \\
       & = \eb\left[I\left(\eta > y_{i}\right)\text{Pr}\left(M_{r} > \eta \mid \mathcal{F}_{\tau_{k}}\right)\right], \label{eqI}
\end{align}
where $\eta=\hat{\mu}_1 + \frac{z}{\sqrt{\gamma(k+1)}}$.
Since $\Theta_{k} \sim \mathcal{N}\left(\hat{\mu}_{1}, \frac{1}{\gamma(k+1)}\right)$, using \hyperref[ineq3]{Inequality 3}, we can write
\begin{align}
\text{Pr}\left(M_{r} > \eta \; \middle| \; \mathcal{F}_{\tau_{k}} = F_{\tau_{k}} \right) \notag 
    & \geq 1 - \left(1-\frac{1}{\sqrt{2\pi}}\frac{z}{z^{2}+1}e^{\frac{-z^2}{2}}\right)^{r} \notag \\
    & \geq 1 - e^{-\frac{r^{1-\frac{\beta}{2}}}{\sqrt{2\beta\pi\ln{r}}}}.
\end{align}
Note that $F_{\tau_{k}}$ is any realization of $\mathcal{F}_{\tau_{k}}$. 
As $\beta \in [1,2)$, there exists a number $h(\beta) \in\mathbb{R}_{>0}$ such that $e^{-\frac{r^{1-\frac{\beta}{2}}}{\sqrt{2\beta\pi\ln{r}}}}\leq \frac{1}{r^2}$ for $r\geq h(\beta)$. Hence, for any $r\geq h(\beta)$ and any $\gamma>0$, 
\begin{equation}\label{maxr}
    \text{Pr}\left( M_{r}   > \eta \; \middle| \; \mathcal{F}_{\tau_{k}} = F_{\tau_{k}} \right) \geq 1 - \frac{1}{r^2}
\end{equation}
On substituting \eqref{maxr} in \eqref{eqI} we get, for any $r\geq h(\beta)$,
\begin{align}
    \text{Pr}(G_k < r) &\geq \eb\left[I\left( \eta \geq y_{i} \right)\left(1-\frac{1}{r^2}\right) \right]\notag \\
    &= \left(1-\frac{1}{r^2}\right)\text{Pr}( \eta \geq y_{i}) \label{eq:ch3}
\end{align}
We will now find a lower bound on $\text{Pr}( \eta \geq y_{i})$.
\begin{multline}
    \text{Pr}(\eta \geq y_{i}) = \text{Pr}\left( \hat{\mu}_{1}+\frac{z}{\sqrt{\gamma(k+1)}} \geq \mu_{1}-\frac{\Delta_{i}}{3}\right)=\\
    \text{Pr}\left( \hat{\mu}_{1}+\frac{1}{k+1}\geq  \mu_{1}-\left(\frac{\Delta_{i}}{3}+\frac{z}{\sqrt{\gamma(k+1)}}-\frac{1}{k+1}\right)\right) \label{eqCH}
\end{multline}
$\frac{1}{k + 1}$ was added to $\hat{\mu}_{1}$ to account for the fact that $\hat{\mu}_1$ is {\em{not}} the average of the past $k$ observations, but it is the sum of the past $k$ observations divided by $(k+1)$. Applying  \hyperref[ineq2]{Inequality 2} to \eqref{eqCH},  
\begin{equation}
    \text{Pr}(\eta \geq y_{i}) \geq 1 - e^{-2\left(\frac{\Delta_{i}\sqrt{(k+1)}}{3}+\frac{z}{\sqrt{\gamma}}-\frac{1}{\sqrt{k+1}}\right)^2}. \label{eq:ch2}
\end{equation}
Substituting \eqref{eq:ch2} back into \eqref{eq:ch3}, for any $r\geq h(\beta)$, 
\begin{align}
    \text{Pr}(G_k < r) &\geq \left(1-\frac{1}{r^2}\right)\left(1 - e^{-2\left(\frac{\Delta_{i}\sqrt{(k+1)}}{3}+\frac{z}{\sqrt{\gamma}}-\frac{1}{\sqrt{k+1}}\right)^2}\right) \notag \\
    & \geq 1 - \frac{1}{r^2} - e^{-2\left(\frac{\Delta_{i}\sqrt{(k+1)}}{3}+\frac{z}{\sqrt{\gamma}}-\frac{1}{\sqrt{k+1}}\right)^2}.
\end{align}
This leads us to 
\begin{multline}
\sum_{k=0}^{4L_{i}(T)-1} \eb[G_k] = \sum_{k=0}^{4L_{i}(T)-1} \sum_{r=0}^{T} \text{Pr}(G_k \geq r) \\ \leq  \sum_{k=0}^{\small{4L_{i}(T)-1}}\sum_{r=0}^{T}   \left( \frac{1}{r^2}  \right. \\ + \left. e^{-2\left(\frac{\Delta_{i}\sqrt{(k+1)}}{3}+\frac{z}{\sqrt{\gamma}}-\frac{1}{\sqrt{k+1}}\right)^2}\right) \label{eq:logt}
\end{multline}
First term on the RHS in (\ref{eq:logt}) can be upper bounded as follows. 
\begin{align}
    \sum_{k=0}^{4L_{i}(T)-1}\left(\sum_{r=0}^{T} \frac{1}{r^2} \right) & 
    \leq \sum_{k=0}^{4L_{i}(T)-1} \left( h(\beta) +  \sum_{r\geq h(\beta)}  \frac{1}{r^2}\right) \notag \\
    & \leq \sum_{k=0}^{4L_{i}(T)-1} \left( h(\beta) +  \zeta(2)\right) \notag \\
    & \leq (h(\beta) +  \zeta(2))4L_{i}(T) \label{eq:smallr},
\end{align}
where $\zeta$ is the Riemann zeta function. Next, we consider the second term on the RHS of (\ref{eq:logt}) and use $k'=k+1$ for convenience.
\begin{align}
   &\sum_{k'=1}^{\small{4L_{i}(T)}}\sum_{r=0}^{T} e^{-2\left(\frac{\Delta_{i}\sqrt{k'}}{3}+\frac{\sqrt{\beta\ln{r}}}{\sqrt{\gamma}}-\frac{1}{\sqrt{k'}}\right)^2} \notag \\
   &=\sum_{r=0}^{T}\sum_{k'=1}^{\small{4L_{i}(T)}} e^{-2\left(\frac{\Delta_{i}\sqrt{k'}}{3}+\sqrt{\frac{\beta\ln{r}}{\gamma}}\right)^2}e^{\frac{-2}{k'}}e^{4\left(\sqrt{\frac{\beta\ln{r}}{k'\gamma}}\right)}e^{\frac{4\Delta_i}{3}} \notag \\
   &\leq\sum_{r=0}^{T}\sum_{k'=1}^{\small{4L_{i}(T)}} e^{-2\left(\frac{\Delta_{i}\sqrt{k'}}{3}+\sqrt{\frac{\beta\ln{r}}{\gamma}}\right)^2}e^{4\left(\sqrt{\frac{\beta\ln{r}}{k'\gamma}}\right)}e^{\frac{4\Delta_i}{3}} \notag \\
   &= \sum_{r=0}^{T}\sum_{k'=1}^{\small{4L_{i}(T)}} e^{\frac{-2\beta \ln{r}}{\gamma}}e^{\frac{-2\Delta_i^2k'}{9}}e^{\frac{-4\Delta_i}{3}\sqrt{\frac{\beta k'\ln{r}}{\gamma}}}e^{4\left(\sqrt{\frac{\beta\ln{r}}{\gamma k'}}\right)}e^{\frac{4\Delta_i}{3}} \notag \\
   & \overset{(a)}{\leq}\sum_{r=0}^{T}e^{\frac{-2\beta\ln{r}}{\gamma}}e^{4\left(\sqrt{\frac{\beta\ln{r}}{\gamma}}\left(1-\frac{\Delta_i}{3}\right)\right)}e^{\frac{4\Delta_i}{3}} \sum_{k'=1}^{\small{4L_{i}(T)}} e^{\frac{-2\Delta_i^2 k'}{9}} \notag \\
   &\overset{}{\leq}\sum_{r=0}^{T} e^{\frac{-2\beta\ln{r}}{\gamma}}e^{4\left(\sqrt{\frac{\beta\ln{r}}{\gamma}}\left(1-\frac{\Delta_i}{3}\right)\right)}e^{\frac{2\Delta_i}{3}}\frac{1}{e^{(\frac{2\Delta_i^2}{9}-1)}} \notag \\
   &\overset{(b)}{=} \sum_{r=0}^{T} \frac{c'}{r^{\frac{2\beta}{\gamma}}}e^{4\left(\sqrt{\frac{\beta\ln{r}}{\gamma}}\left(1-\frac{\Delta_i}{3}\right)\right)}
\end{align} 
$(a)$ is due to the fact that $\max\left(e^{\frac{-4\Delta_i}{3}\left(\sqrt{\frac{\beta k'\ln{r}}{\gamma}}\right)}e^{4\left(\sqrt{\frac{\beta\ln{r}}{\gamma k'}}\right)}\right) = e^{4\left(\sqrt{\frac{\beta\ln{r}}{\gamma}}\left(1-\frac{\Delta_i}{3}\right)\right)}$ at $k'=1$.
In $(b)$, $c'=e^{\frac{4\Delta_i}{3}}/(e^{\frac{2\Delta_i^2}{9}-1}) $. For any $\epsilon > 0$, there exists a number $g(\epsilon$) such that $\frac{e^{4\sqrt{\frac{\ln{r}}{\gamma}}\left(1-\frac{\Delta_i}{3}\right)}}{r^{\frac{2\beta}{\gamma}}} \leq \frac{1}{r^{\frac{2\beta}{\gamma}-\epsilon}}$ for $r\geq g(\epsilon)$. Hence, for $\beta\in[1,2)$, $\gamma>0$, $\epsilon>0$ and $r\geq g(\epsilon)$, 
\begin{equation} \label{eq:mid}
    \sum_{k'=1}^{\small{4L_{i}(T)}}\sum_{r=0}^{T} e^{-2\left(\frac{\Delta_{i}\sqrt{k'}}{3}+\frac{\sqrt{\beta\ln{r}}}{\sqrt{\gamma}}-\frac{1}{\sqrt{k'}}\right)^2} \leq \sum_{r=0}^{T}\frac{c'}{r^{\frac{2\beta}{\gamma}-\epsilon}}
\end{equation}
We will analyze \eqref{eq:mid} separately for $\gamma\in(0,4)$ and $\gamma\geq 4$.

For any value of $\gamma\in(0,4)$, we choose $\beta\in[1,2)$ such that $\gamma<2\beta$. Then, we select $\epsilon>0$ to have $\frac{2\beta}{\gamma}-\epsilon>1$. Thus, for $\gamma\in(0,4)$, \eqref{eq:mid} can be further simplified as,
\begin{align}
    \sum_{k'=1}^{\small{4L_{i}(T)}}\sum_{r=0}^{T} e^{-2\left(\frac{\Delta_{i}\sqrt{k'}}{3}+\frac{\sqrt{\beta\ln{r}}}{\sqrt{\gamma}}-\frac{1}{\sqrt{k'}}\right)^2} 
    \leq \sum_{r=0}^{T}\frac{c'}{r^{\frac{2\beta}{\gamma}-\epsilon}} \notag \\
    \leq c'g(\epsilon)+\sum_{r\geq c'g(\epsilon)}\frac{c'}{r^{\frac{2\beta}{\gamma}-\epsilon}} \notag \\
    \leq c' \left(g(\epsilon) + \zeta\left(\frac{2\beta}{\gamma}-\epsilon\right)\right)\label{eq:smalla}
\end{align}

Since $\frac{2\beta}{\gamma}-\epsilon>1$ and $\zeta$ is the Riemann zeta function,  $\zeta\left(\frac{2\beta}{\gamma}-\epsilon\right)$ is a finite number.

On the other hand, for $\gamma\geq4$, $\frac{2\beta}{\gamma}-\epsilon<1$ for any choice of $\beta$ and $\epsilon$. If we fix $\beta\in[1,2)$ and $\epsilon>0$ such that $\frac{2\beta}{\gamma}-\epsilon > 0$ \eqref{eq:mid} results in,  

\begin{align}
    \label{eq:biga}
     \sum_{k'=1}^{\small{4L_{i}(T)}}\sum_{r=0}^{T}& e^{-2\left(\frac{\Delta_{i}\sqrt{k'}}{3}+\frac{\sqrt{\beta\ln{r}}}{\sqrt{\gamma}}-\frac{1}{\sqrt{k'}}\right)^2} \notag \\ & \leq \sum_{r=0}^{T}\frac{c'}{r^{\frac{2\beta}{\gamma}-\epsilon}} \notag \\
     & \leq c'\left( g(\epsilon)+\sum_{r\geq1}\frac{c'}{r^{\frac{2\beta}{\gamma}-\epsilon}} \right)\notag \\
     & \overset{(a)}{\leq} c'\left( g(\epsilon) + 1 + \frac{T^{1+\epsilon-\frac{2\beta}{\gamma}}-1 }{1+\epsilon-\frac{2\beta}{\gamma}}\right)
\end{align}
The inequality $(a)$ in \eqref{eq:biga} follows from \hyperref[ineq5]{Inequality 5}. On substituting \eqref{eq:smalla}, \eqref{eq:biga} and \eqref{eq:smallr} back into \eqref{eq:logt} gives us the bound for Sum (\ref{eq:sums}a), the first term on the RHS of \eqref{eq:sums}.

\begin{multline}
\sum_{k=0}^{4L_{i}(T)-1} \eb[G_k] \leq \\ \left\{ \begin{array}{l}
    H(\beta) L_i(T)+ c' \left (g(\epsilon)+\zeta(\frac{2\beta}{\gamma}-\epsilon)\right)~ {\text{for}}~\gamma\in(0,4),\\
    H(\beta)L_i(T)+c' \left( g(\epsilon) + \frac{T^{1+\epsilon-\frac{2\beta}{\gamma}}-1 }{1+\epsilon-\frac{2\beta}{\gamma}}\right)~{\text{for}}~\gamma \geq 4,
\end{array}
\right.
\label{eq:suma}
\end{multline}
where, $H(\beta)=4(h(\beta) +  \zeta(2))$.

Next, we bound Sum (\ref{eq:sums}b), second term on the RHS of \eqref{eq:sums}) where the index of summation  $k\geq4L_i(T)$. We will start by defining $A_{t-1}$ as the event in which $\hat{\mu}_1(t)-\frac{\Delta_i}{6}>y_i$ and 
use the notation $\mathcal{F}_{t-1}|_{A_{t-1}}$ to indicate random variable $\mathcal{F}_{t-1}$ conditioned on $A_{t-1}$ being true. Then,
\begin{align}
    \eb\left[\frac{1}{p_{i,\tau_{k}+1}}\right] &=  \eb\left[\frac{1}{\text{Pr}(\Theta_k>y_i|\mathcal{F}_{\tau_k})}\right] \notag \\
    & \leq \eb\left[\frac{1}{\text{Pr}\left(\Theta_k>y_i\Big|\mathcal{F}_{\tau_k}|_{A_{\tau_k}}\right)\text{Pr}(A_{\tau_k})}\right]  \label{eq:main}
\end{align}
We now bound $\text{Pr}(\Theta_k>y_i|\mathcal{F}_{\tau_k}|_{A_{\tau_k}})$ and $\text{Pr}(A_{\tau_k})$.
\begin{align}
   \text{Pr}\left(\Theta_k>y_i \; \middle| \;\mathcal{F}_{\tau_k}|_{A_{\tau_k}}\right) &\geq \text{Pr}\left(\Theta_k>\hat{\mu}_1-\frac{\Delta_i}{6} \; \middle|\; \mathcal{F}_{\tau_k}|_{A_{\tau_k}} \right) \notag \\
   &\overset{(a)}{\geq} 1-e^{-\gamma(k+1)\Delta_i^2/72} \notag \\
   &\overset{(b)}{\geq} 1-e^{-\gamma(4L_i(T))\Delta_i^2/72} \notag \\
   &\geq 1- \frac{1}{T\Delta_i^2} \label{eq:Y}
\end{align}
In the above, $(a)$ follows from \hyperref[ineq2]{Inequality 2} with $z=\sqrt{\gamma(k+1)}\Delta_i/6$ and $(b)$ is due to the fact that $k\geq4L_i(T)$. Note that we can use \hyperref[ineq2]{Inequality 2} here because we assume that the reward distribution has a finite support over $[0,1]$.
 
Observe that for any $t\geq\tau_k+1$, we have $k_1(t)\geq k\geq4L_i(T)$, and, using \hyperref[ineq2]{Inequality 2}, we obtain,
\begin{align}
\text{Pr}(A_{\tau_k}) = \text{Pr}\left(\hat{\mu}_1(t) > \mu_1 - \frac{\Delta_i}{6}\right) &\geq 1- e^{-2\gamma k_1(t)\Delta_i^2/36} \notag \\
&\geq 1 - \frac{1}{T\Delta_i^2} \label{eq:A}
\end{align}
Substituting \eqref{eq:Y} and \eqref{eq:A} into \eqref{eq:main}, for \small{$k\geq4L_i(T)$},
\begin{align}
    \eb\left[\frac{1}{p_{i,\tau_{k}+1}}\right] -1 &\leq \frac{1}{\left(1- \frac{1}{T\Delta_i^2}\right)^2}-1 \notag\\
    &\leq \frac{4}{T\Delta_i^2} \label{eq:highk}
\end{align}
{\bf{For any}} $\gamma >0$, using \eqref{eq:highk}, we get the following bound on Sum (\ref{eq:sums}b).  
\begin{align}
    \sum_{\small{4L_{i}(T)}}^{T-1} \eb[G_k] &\leq \sum_{\small{4L_{i}(T)}}^{T-1} \frac{4}{T\Delta_i^2} \notag \\
    & \leq \frac{4}{\Delta_i^2} \label{eq:sumb}
\end{align}
Combining the results from \eqref{eq:suma}, \eqref{eq:sumb}, \eqref{eq:sums} and \eqref{eqn:termA1} completes the proof of Lemma \ref{lem:termA}.

\begin{IEEEbiography}[{\includegraphics[width=1in,height=1.25in,clip,keepaspectratio]{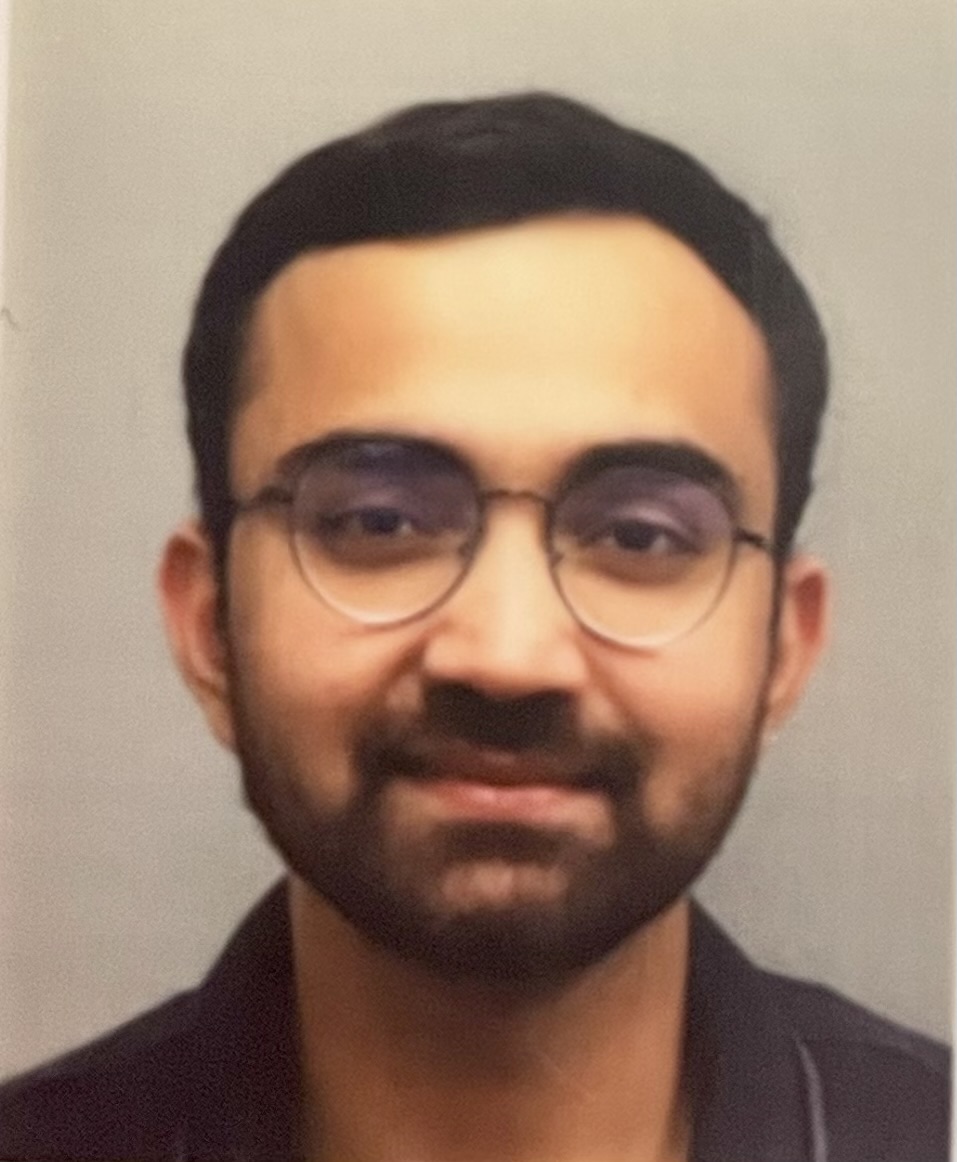}}]%
{Kartik Anand Pant}{\space} was born in Bhopal, Madhya Pradesh, India. He received his B.Tech. degree in electronics engineering from IIT (BHU) Varanasi, India, in 2018. He was a Scientist at Indian Space Research Organization, India, from 2018 to 2021.

He is currently a Master's student and a Graduate Research Assistant in Aeronautical and Astronautical Engineering at Purdue University. His research interests include Robotics and Control theory. He is a recipient of the Chintakindi Amba Rao Fellowship at Purdue University in 2021.
\end{IEEEbiography}

\begin{IEEEbiography}[{\includegraphics[width=1in,height=1.25in,clip,keepaspectratio]{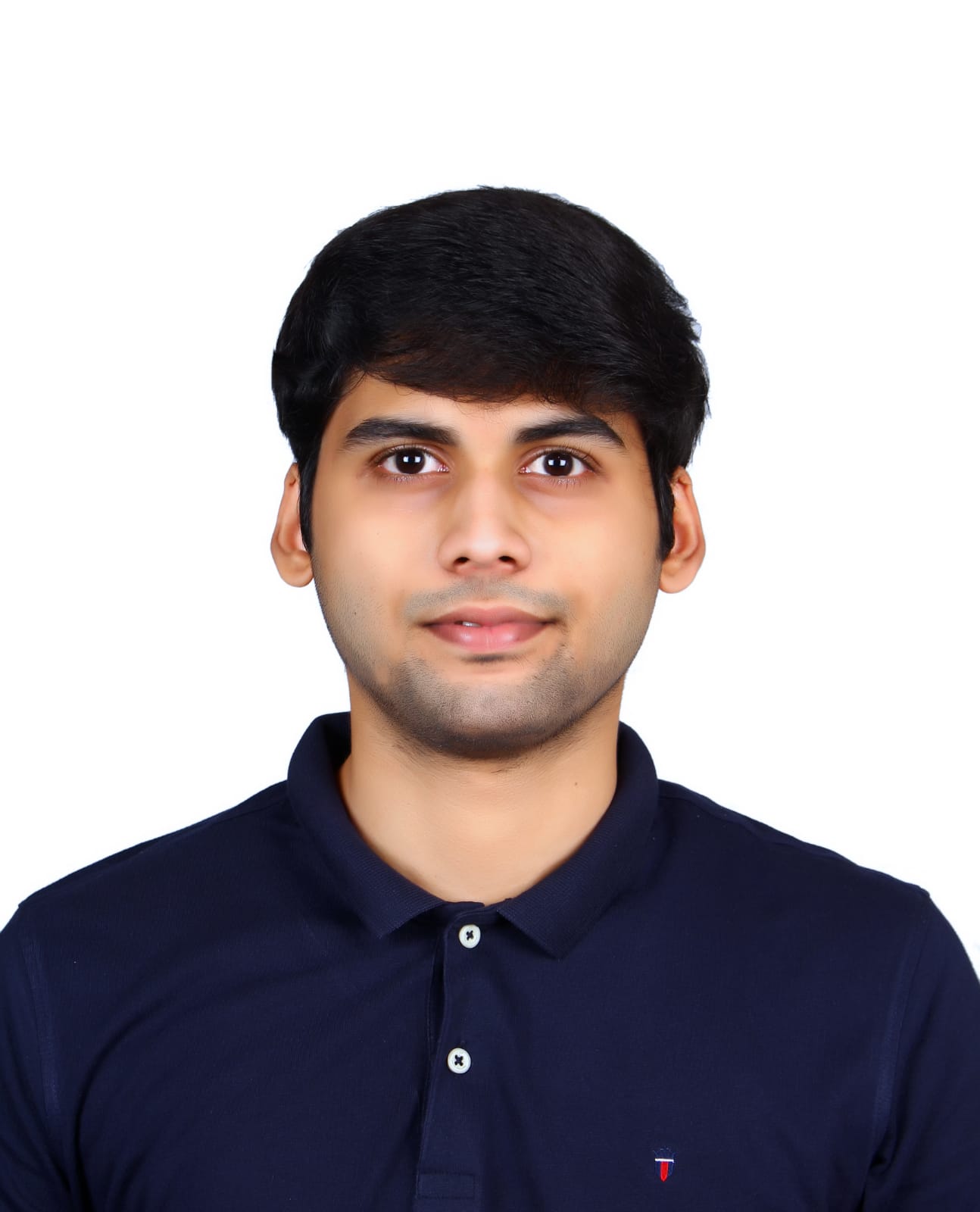}}]%
{Amod Hegde}{\space}was born in Sirsi, Karnataka,
India, in 1996. He received the B.Tech. degree in electronics engineering from IIT (BHU) Varanasi, India, in 2018. 

He is currently a Master's student in computational social science at Stanford University. His research interests include optimization techniques/theory and machine learning. He was a recipient of the Director’s Gold Medal at IIT BHU Varanasi, for outstanding all-round performance in the graduating class
\end{IEEEbiography}

\begin{IEEEbiography}[{\includegraphics[width=1in,height=1.25in,clip,keepaspectratio]{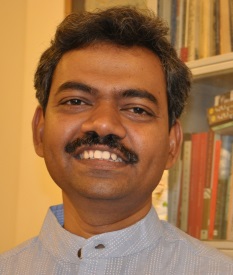}}]%
{K. V. Srinivas}{\space}was born in Vijayawada, India. He received the B.E. degree in electronics and communications engineering from the Andhra University College of Engineering, Vishakhapatnam, India, in June 1996, the M.Tech. degree from the Indian Institute of Technology, Kanpur, India, in 1998, and the Ph.D degree from the Indian Institute of Technology Madras, Chennai, India, in 2009, both in electrical engineering. 

He was a Postdoctoral Fellow at the Department of Electrical and Computer Engineering, University of Toronto, from March 2009 to October 2011. From January 2015 to December 2018, he was a Faculty member at the Indian Institute of Technology (BHU), Varanasi, India. His past industry experience includes working at the Indian Space Research Organisation, Samsung Electronics and Nokia Networks and Ericsson. In 2022 he joined Motorola Mobility/Lenovo as a researcher working on 5G NR standardization activities. His research interests include wireless communications, with emphasis on physical and MAC layer algorithms, and machine learning.
\end{IEEEbiography}

\end{document}